\documentclass[table]{hfstyle/hf}

\usepackage{graphicx}
\usepackage{booktabs}
\usepackage{enumitem}
\usepackage{graphicx}
\usepackage{subcaption}
\usepackage{rotating} 

\newcommand{\huggingface}{\raisebox{-1.5pt}{\includegraphics[height=1.05em]{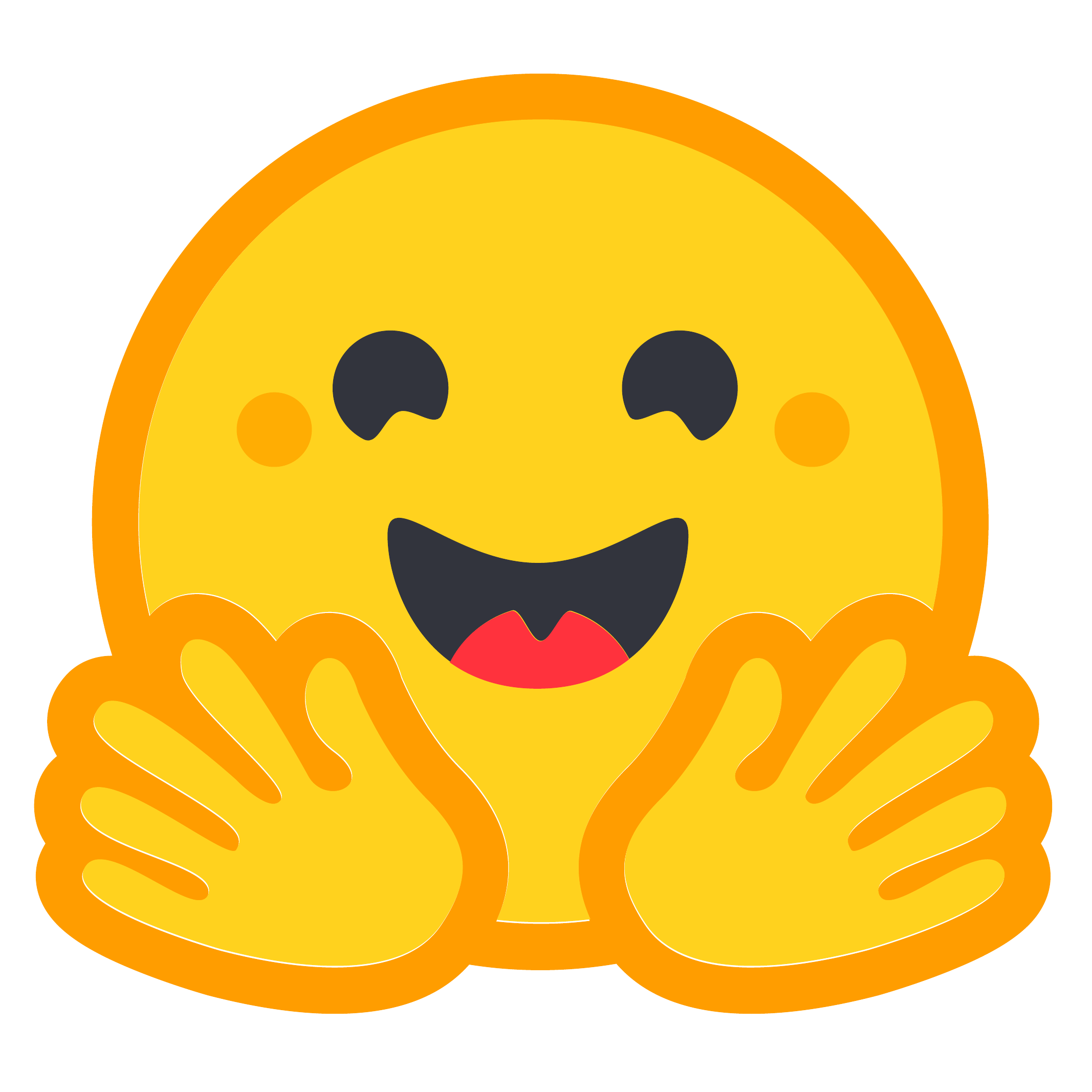}}}
\newcommand{\coreContrib}{\raisebox{.33em}{\hspace{.05em}\includegraphics[height=.5em]{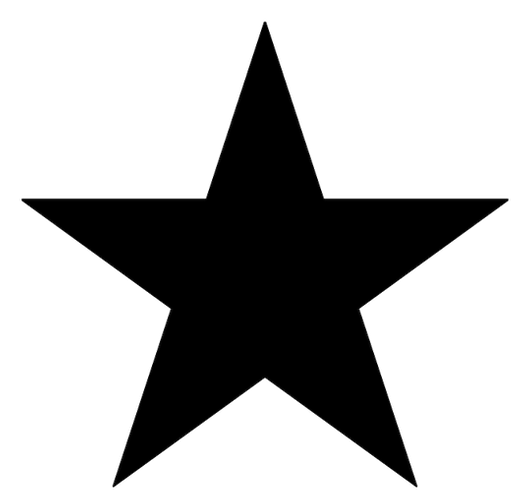}}}
\newcommand{\hf}{\raisebox{.28em}{\hspace{.05em}\includegraphics[height=.65em]{logos/hf.pdf}}\hspace{.05em}}
\newcommand{\pollen}{\raisebox{.28em}{\hspace{.05em}\includegraphics[height=.60em]{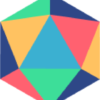}}\hspace{.05em}}

\newcommand{\spaces}{\raisebox{-1.5pt}{\includegraphics[height=1.05em]{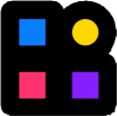}}}
\newcommand{\hfdataset}{\raisebox{-1.5pt}{\includegraphics[height=1.05em]{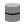}}}
\newcommand{\gitrepo}{\raisebox{-1.5pt}{\includegraphics[height=1.05em]{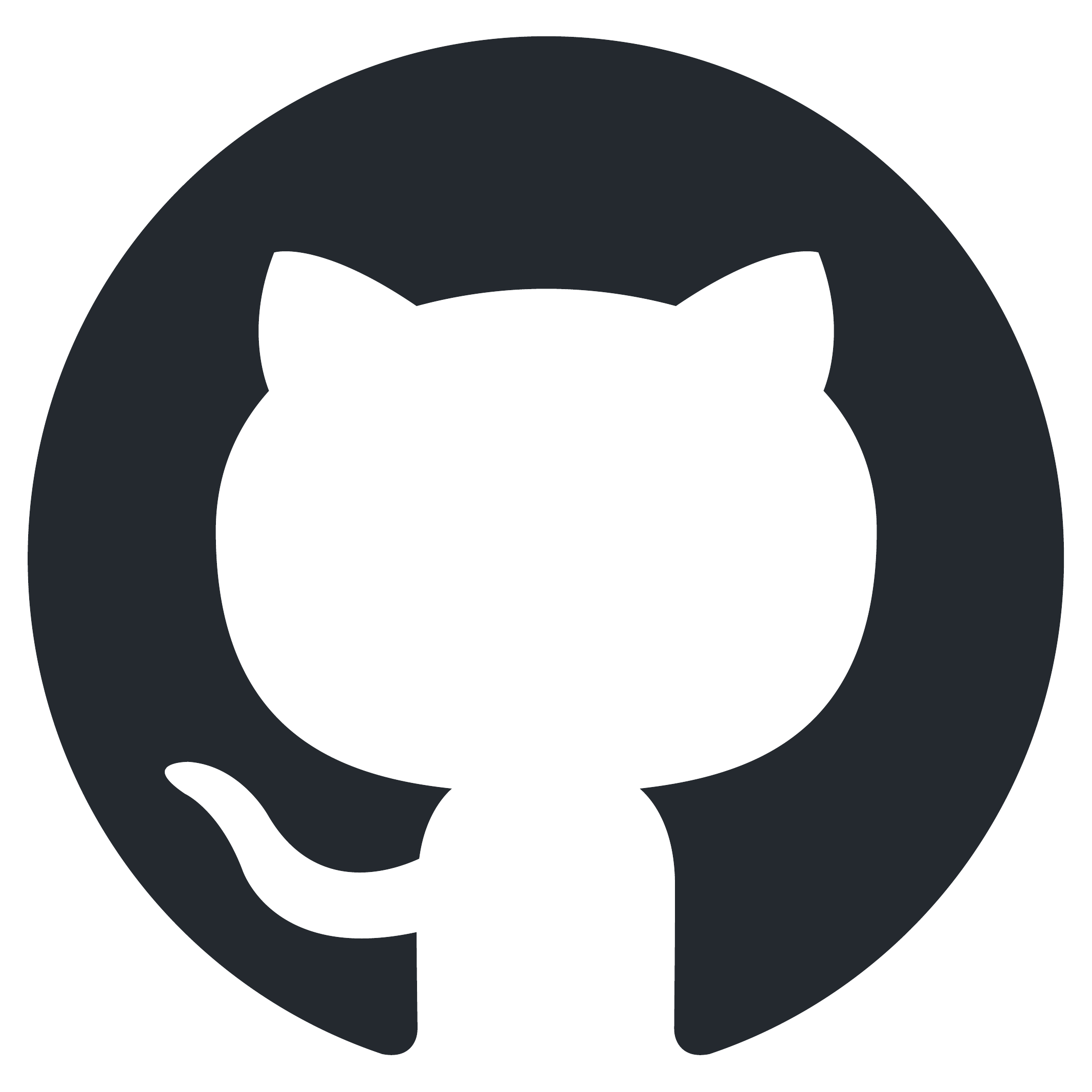}}}

\newtcolorbox{academicNote}{
  colback=gray!5!white,      
  colframe=gray!50!black,    
  fonttitle=\bfseries,
  title=Note,
  arc=1mm,                   
  left=8pt, right=8pt, top=4pt, bottom=4pt
}

\microtypesetup{expansion=false}

\setlist[itemize]{leftmargin=*,itemsep=0.2em,parsep=0.2em,topsep=0.3em}

\title{The Sustainability Gap in Robotics:\\ A Large-Scale Survey of Sustainability\\Awareness in 50,000 Research Articles}

\authorOne[]{Antun Skuric\hf \pollen\coreContrib}
\authorOne[]{Leandro Von Werra\hf}
\authorOne[]{Thomas Wolf\hf}
\affiliation[]{\hf Hugging Face, \pollen Pollen Robotics, \coreContrib Corresponding Author}

\setmainfigure{
    \begin{figure}[!h]
        \centering
        \includegraphics[width=0.8\textwidth, trim={0 1cm 0 1cm},clip]{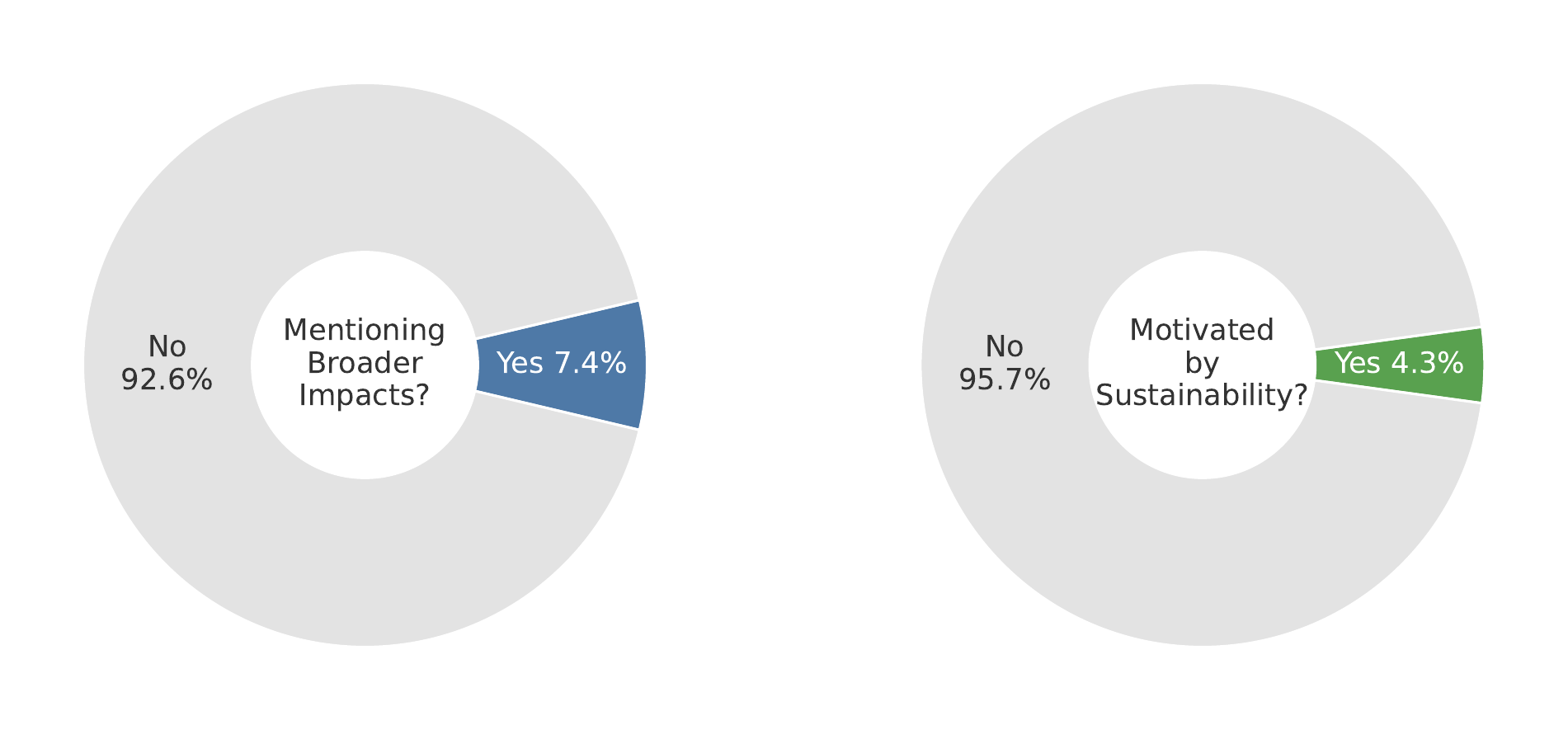}
        \caption{\textbf{The Sustainability Gap}: Out of 50,000 analyzed papers, only 7\% mention broader societal 
        or ecological impacts, and only about 4\% are explicitly
        motivated by sustainability-related goals.}
                \label{fig:teaser}
    \end{figure}
}

\abstract{
We present a large-scale survey of sustainability communication and 
motivation in robotics research. Our analysis covers nearly 50,000 
open-access papers from arXiv's cs.RO category published between 2015 
and early 2026. In this study, we quantify how often papers mention social,
ecological, and sustainability impacts, and we analyse their alignment 
with the UN Sustainable Development Goals (SDGs). 

\medskip

The results reveal a  persistent gap between the field's potential 
and its stated intent. While a large fraction of robotics papers can be 
mapped to SDG-relevant
domains, explicit sustainability motivation remains remarkably low. 
Specifically, mentions of sustainability-related impacts are typically 
below 2\%, explicit SDG references stay below 0.1\%, and the proportion
of sustainability-motivated papers remains below 5\%. These trends 
suggest that while the field of robotics is advancing rapidly, 
sustainability is not yet a standard part of research framing. 

\medskip

We conclude by proposing concrete actions for researchers, conferences, 
and institutions to close these awareness and motivation gaps, 
supporting a shift toward more intentional and responsible innovation.
}

\setmaintable{
\begin{table}[!h]
    \centering
    \begin{tabular}{l l}
        \spaces~{\sans{\textbf{Demo}}} & \href{https://huggingface.co/spaces/sustainable-robotics/paper-sustainability-assessment}{\texttt{sustainable-robotics/paper-sustainability-assessment}} \\
        \huggingface~{\sans \textbf{Blog}} & \href{https://huggingface.co/spaces/sustainable-robotics/sustainability-in-robotics}{\texttt{sustainable-robotics/sustainability-in-robotics}} \\
        \hfdataset~{\sans \textbf{Dataset}} & \href{https://huggingface.co/datasets/sustainable-robotics/robotics-arxiv-sustainability-classification}{\texttt{sustainable-robotics/robotics-arxiv-sustainability-classification}} \\
        \gitrepo~{\sans \textbf{Github Repo}} & \href{https://github.com/huggingface/arxiv-robotics-sustainability-classification}{\texttt{huggingface/arxiv-robotics-sustainability-classification}} \\
    \end{tabular}
\end{table}
}

\usepackage[most]{tcolorbox}

\begin{document}

\maketitle
\vspace{-1cm}
\section{Introduction}

Robotics is gradually expanding beyond highly specialized industrial applications toward more versatile, 
general-purpose systems. This evolution, initially driven by the adoption of human-centered collaborative 
robots, is now increasingly focused on humanoid platforms. Through decades of continuous research and 
innovation, these systems are achieving impressive levels of mobility\footnote{2026 Spring Festival Gala by China Media Group - 
\href{https://www.youtube.com/watch?v=mUmlv814aJo}{Video from CGTN}} and dexterity, with highly 
capable platforms now beginning to enter the commercial market\footnote{\href{https://www.unitree.com/g1}{https://www.unitree.com/g1}}.

Complementing these physical advancements, recent progress in artificial intelligence (AI) has significantly enhanced
the \textit{cognitive capabilities} of robots~\citep{WANG2025}. The integration of large language models (LLMs) and other 
AI techniques enables these systems to understand and interact with the world in increasingly sophisticated 
ways. While the field must still overcome substantial technical challenges to achieve 
robust, real-world autonomy~\citep{riener2023robots}, these combined physical and cognitive improvements 
allow us to envision robots performing vital societal roles.

Traditionally, the primary applications of robotics have focused on substituting human labor, 
mitigating exposure to hazardous environments, reducing physical strain from repetitive tasks, 
and optimizing resource management through high-precision automation. Nowadays, the field is 
increasingly shifting toward collaborative assistance, entering the realm of daily human life. 
Applications such as providing essential care for the elderly and assisting with everyday 
household tasks are increasingly becoming within reach, with several robotics companies already beginning to 
commercialize these services\footnote{\href{https://www.1x.tech/neo}{1X Neo - Home Robot}}.

The combination of AI and robotics seems to open unprecedented possibilities for the future of the field~\citep{Takur2025}.
However, as the scope and scale of these robotic applications grow, so does the imperative 
to critically evaluate their broader societal and environmental footprints.

\subsection{The Reality of Planetary Boundaries}

As robotics transitions toward ubiquitous societal deployment, its development is frequently driven 
by an unspoken rule: \textit{"if it can exist, it should exist"}. This mindset, often compounded 
by competitive pressures (\textit{"if we don't build it, someone else will"}), has accelerated innovation but obscured a fundamental physical constraint: 
we live on a finite planet with limited resources~\citep{meadows1972limits}. 
The paradigm of unconstrained technological
 progress now collides with an unprecedented global ecological crisis.

\begin{figure}[!htb]
    \centering
    \includegraphics[width=0.7\linewidth]{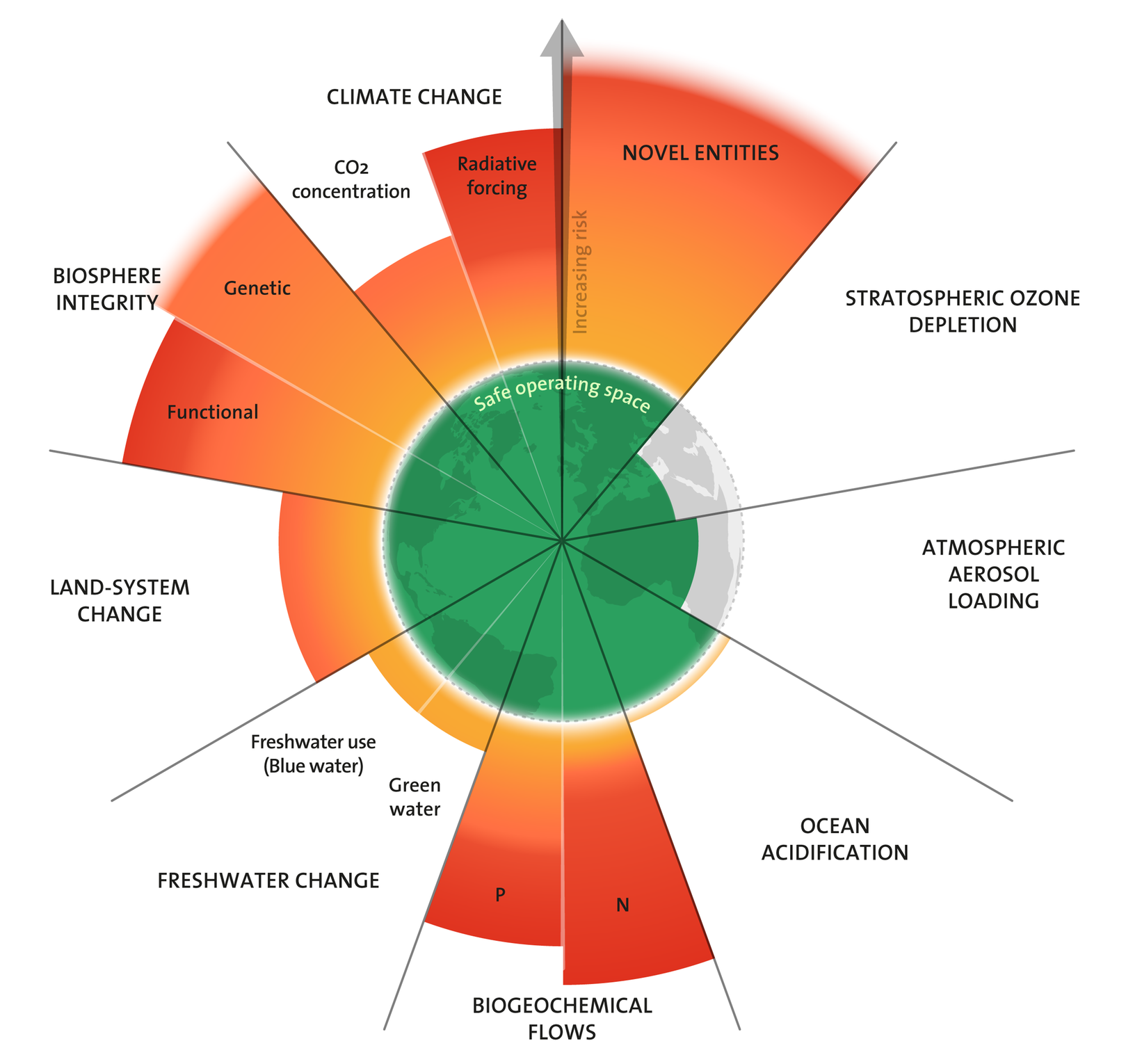}
    \caption{2025 update of planetary boundaries. Figure adapted from publicly 
    released material by the Stockholm Resilience Centre and collaborators~\citep{kitzmann2025planetary}.}
    \label{fig:planetary_boundaries}
\end{figure}

Scientific evidence, including the \href{https://www.ipcc.ch/report/ar6/syr/}{latest IPCC reports}, 
confirms that our current trajectory is environmentally unsustainable. This crisis is best 
quantified by the \textit{"planetary boundaries"} framework~\citep{rockstrom2009safe}, which defines a \textit{"safe operating space"} for 
humanity across nine critical Earth systems. Exceeding these boundaries increases the risk 
of triggering irreversible \textbf{tipping points} that threaten the foundation of life on Earth.

According to the latest Planetary Health Check 2025~\citep{kitzmann2025planetary}, humanity has already surpassed seven 
of these nine limits, pushing us into a high-risk zone across multiple fronts:\footnote{Read more about the evolution of the planetary boundaries at the 
\href{https://www.stockholmresilience.org/research/planetary-boundaries.html}{Stockholm Resilience Centre}.}

\begin{itemize}
    \item \textbf{Climate Change:} Atmospheric carbon concentrations have long since
     passed the safe threshold~\citep{Richardson2023}, and the average global temperature exceeded 1.5°C in 2024 (the target set by the \href{https://unfccc.int/process-and-meetings/the-paris-agreement}{Paris Agreement}). 
    \item \textbf{Biosphere Integrity:} Biodiversity is being lost at an alarming rate~\citep{johnson2017biodiversity}.
    \item \textbf{Land-System Change:} Deforestation and aggressive land use are 
    destabilizing local and global climates~\citep{chakravarty2012deforestation,smith2023tropical}.
    \item \textbf{Novel Entities:} Chemical pollution and microplastics have become
     ubiquitous, found in the most remote corners of the Earth~\citep{law2014microplastics,world2019microplastics} 
     and even within the human bloodstream~\citep{cox2019human}.
\end{itemize}

The fact that we have exceeded the majority of our planet's safe operating limits 
is a clear indicator that we need to rethink our approach to innovation. Achieving true \textit{sustainability}  
requires more than simply reducing carbon emissions; it demands a fundamental reimagining of 
how we design, deploy, and scale technologies. It requires simultaneous action on 
multiple fronts: managing resource utilization, reducing inequalities, ensuring 
peace, and aiming for equitable economic prosperity. Ultimately, it is about 
creating a stable paradigm that balances our current technological ambitions
with the needs of future generations and the planet itself~\citep{brundtland1987our}.

\subsection{The 2030 Agenda for Sustainable Development}

\begin{figure}[!htb]
    \centering
    \includegraphics[width=0.65\linewidth]{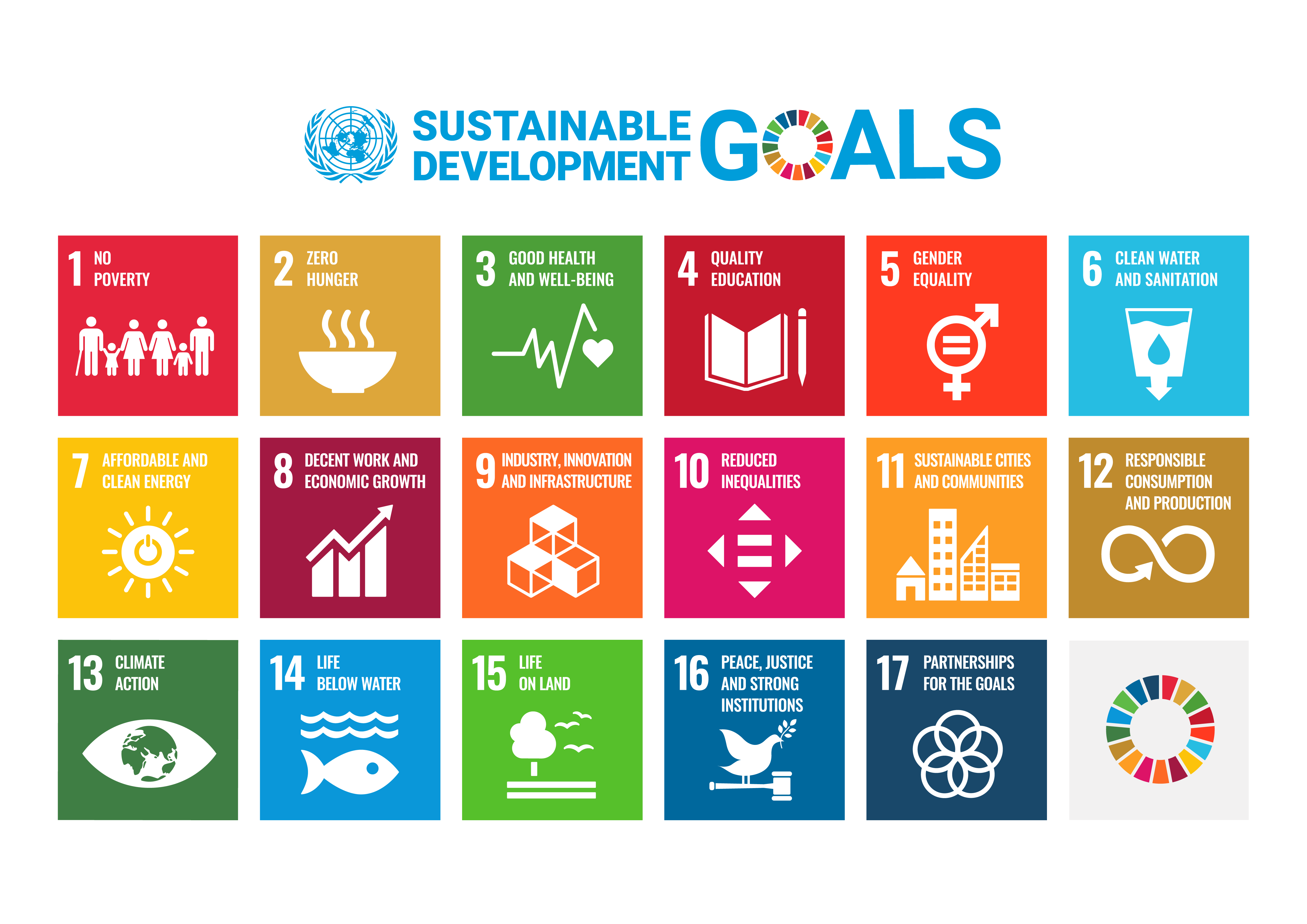}
    \caption{The 17 UN Sustainable Development Goals (SDGs), used as the reference framework for our motivation and alignment analysis.}
    \label{fig:sdg_poster}
\end{figure}

The year 2015 marked a pivotal moment for global action. While many remember it as the year 
of the \href{https://unfccc.int/process-and-meetings/the-paris-agreement}{Paris Agreement}, 
it was also when all United Nations Member States adopted a shared blueprint for a sustainable 
transition called the \href{https://sdgs.un.org/2030agenda}{2030 Agenda for Sustainable Development}.

\begin{quote}
\textit{"The 2030 Agenda for Sustainable Development, adopted by all United Nations Member 
States in 2015, provides a shared blueprint for peace and prosperity for people and the 
planet, now and into the future."}
\end{quote}     

At the heart of this blueprint are the \textbf{17 Sustainable Development Goals (SDGs)}\footnote{Read more about the \href{https://sdgs.un.org/goals}{UN's SDGs}.}. 
They represent one of the most comprehensive frameworks of the actions necessary to accomplish 
a global transition within a 15-year window (until 2030). They cover a wide range of interconnected 
issues, from poverty and hunger to climate action and responsible consumption. The goals are further 
divided into 169 targets representing relatively concrete and measurable milestones that humanity 
must aim to achieve by 2030.

Each year, the UN produces \href{https://unstats.un.org/sdgs/report/2025/}{a progress report} tracking 
collective progress toward these targets, providing a data-driven mirror of where we stand. Additionally, 
every four years, a comprehensive \href{https://sdgs.un.org/gsdr}{Global Sustainable Development Report (GSDR)} is published. 

The common theme in every report is clear: \textbf{we are falling behind}, and the call for urgent, transformative action grows 
louder every year\footnote{See the latest GSDR report from 2023: \href{https://sdgs.un.org/gsdr/gsdr2023}{Times of Crisis}.}.

\subsection{The Intersection of Robotics and Global Sustainability}

Analogous to AI~\citep{vinuesa2020role, Sasha2023tackling}, researchers increasingly recognize that robotics acts as 
a \textbf{double-edged tool}~\citep{guenat2022meeting, bugmann2011role}. 

On one hand, it possesses the potential to act as one of the primary engines for the transition to a sustainable 
world. Robotics can drive progress across numerous domains, from improving the efficiency of industrial
processes and enabling advanced environmental monitoring, to enhancing human well-being through 
medical applications. On the other hand, if we do not explicitly focus on sustainability, 
robotics can become part of the problem. Primary concerns include contributing to waste 
creation, accelerating energy and resource depletion, and amplifying social inequities.
More precisely, recent analysis suggests that while robotics has the potential to enable 46\% 
of all SDG targets, it could also inhibit 19\% of them~\citep{haidegger2023robotics}.

To ensure that robotics follows a sustainable development trajectory, a concerted effort is required to raise awareness, build 
robust tools, and share knowledge within and across the community. 
Recent institutional moves across industry and academia reflect this necessary shift:

\begin{itemize}
\item \textbf{In Industry:} The International Federation of Robotics (IFR) has published a 
detailed list of \href{https://ifr.org/sustainability#}{proposals}\footnote{See more about IFR proposals at \href{https://ifr.org/sustainability}{https://ifr.org/sustainability}} outlining how robotics can contribute to specific SDGs.
\item \textbf{In Academia:} The IEEE Robotics and Automation Society (RAS), a leading global 
robotics association, recently established a 
\href{https://www.ieee-ras.org/about-ras/governance/standing-committees/sustainability-and-climate-change/}{Sustainability Committee}\footnote{See more about RAS Sustainability Committee at \href{https://www.ieee-ras.org/about-ras/governance/standing-committees/sustainability-and-climate-change/}{https://www.ieee-ras.org/}} 
to promote these values in research and practice.
\end{itemize}

Building a sustainable future requires considerable effort on many fronts; however, 
to understand the true trajectory of the field, we must first evaluate its foundational output: 
\textbf{academic research}. Research serves as the foundation for the future of the industry; the ideas, 
priorities, and ethical frameworks established in today's papers become the industrial standards of tomorrow.

While a growing body of literature explores topics from energy efficiency to social 
impact~\citep{guenat2022meeting, haidegger2023robotics}, it remains unclear how deeply the broader research 
community has internalized the SDGs, or to what extent it actively attempts to align with them. 

To address this critical gap, this article presents a large-scale systematic study of the scientific 
communication seeking to answer a fundamental question:

\vspace{0.5em}
\noindent \textbf{In 2015, the world agreed on a plan for the future. How has robotics research,
 with its potentially massive future impact, aligned with this goal?}
\section{Analysis of the Current State: A Large-Scale Survey}
\label{sec:methodology}

The primary objective of this study is to understand how the robotics research community communicates its work. 
Specifically, the aim is to systematically quantify: 1) the awareness of the community about its broader impacts 
and 2) the overall implication in tackling the challenges of the global transition to a more sustainable world. 
More precisely, we seek to address two central research questions:
\begin{enumerate}
    \item \textbf{To what extent do authors explicitly communicate the potential broader impacts of their research 
    (e.g., social, sustainability, and ecological consequences)?}
    \item \textbf{What proportion of the literature is explicitly motivated by sustainability-oriented objectives?}
\end{enumerate}

Furthermore, we analyze the temporal evolution of these trends to determine whether field-wide awareness and 
intentionality are improving over time. In order to evaluate the alignment of the research topics
with the sustainability goals in this study, we used the UN's SDG framework as a widely accepted and 
relatively complete set of actions and goals.  

While previous studies have explored the intersection of robotics and sustainability through manual expert 
assessment on a localized scale~\citep{guenat2022meeting}, there remains a critical lack of large-scale analyses 
assessing the field's global tendencies and overall trajectory.

To bridge this gap, we conducted a comprehensive review of the open-access robotics literature available on arXiv 
(categorized under the \texttt{cs.RO}\footnote{See the latest robotics articles on arXiv in the \href{https://arxiv.org/list/cs.RO/recent}{\texttt{cs.RO} category}.} taxonomy), covering the period from 2015 to early 2026. This dataset contains 
about 50,000 research articles, providing an extensive view of the field's evolution since the adoption of the UN SDGs in 2015.

It is important to note that the arXiv database does not contain all robotics publications. 
Certain closed-access databases, such as 
IEEE Xplore and ScienceDirect, contain additional manuscripts not deposited on arXiv. However, as the largest 
\textbf{open-access} repository for robotics research, arXiv provides a sufficiently large and representative 
sample to robustly analyze broad trends in the field's scientific communication.

\subsection{The Analysis Approach}

Addressing these research questions necessitated a methodology combining \textit{quantitative keyword analysis} (Question 1: Impact mentions)
with \textit{qualitative content analysis} (Question 2: Paper motivations). Given the prohibitive time and resources required for manual human 
annotation of this corpus, we developed a high-throughput computational pipeline utilizing the \textbf{DeepSeek-V3}
Large Language Model (LLM). This specific model was selected for its \textbf{open-weight} availability, 
advanced reasoning capabilities, and relatively large context window, which permitted the processing of the 
full text of each manuscript.

Recent studies suggest that the zero-shot classification methods, providing the LLM with a well-structured prompt as well as a complete taxonomy of labels,
are effective tools for classifying text~\citep{wang2023large,vajjala2025text}. Such methods have been used for medical data~\citep{Mahadik2025}, 
sentiment analysis~\citep{Wang2024class} but also news and review classification~\citep{kostina2025large}. Additionally, 
LLMs natural language understanding capabilities have been shown to be effective for extracting climate-related 
information from large documents and reports~\citep{luccioni2020analyzing}.

In this study, the model was tasked with analyzing each paper against two specific criteria: 1) identify presence of impact-related keywords,
2) classify the research based on its alignment with the SDGs.
The classification was framed as a zero-shot approach utilizing a double (system + per-paper) prompt structure. 

The system prompt was engineered to set the tone of the classification: mitigate optimistic or pessimistic biases, produce
reproducible categorization and avoid hallucinations. To ensure consistent categorization the LLM 
was provided with the full taxonomy of labels, by including the official text of the UN SDG targets\footnote{Find the source document used in the prompts here: 
\href{https://huggingface.co/spaces/sustainable-robotics/sustainability-in-robotics/blob/main/app/src/content/assets/docs/un_sdgs.txt}{Download UN SDG Targets Text}}.
Additionally, the requirements of the classification were provided to the model with the 
information about the expected response format, classification structure and the reasoning guidelines. 

Subsequently, for each classified paper, the full text of each PDF was extracted and sent through a
separate prompt to the LLM. The prompt enforced a standardized classification schema detailing the 
paper's explicit motivation, implicit SDG alignment, and its communication of broader impacts. 
Crucially, the model was required to provide a concise justification for each of the 
classification points, including direct quotes from the manuscript, to verify that conclusions 
were anchored in the source text.

More details about the prompt architecture and response examples can be found in the \Cref{sec:appendix}.

\paragraph{Note:} The objective of this methodology is not to provide an accurate judgment of any 
individual paper, but rather to empirically demonstrate overarching trends regarding how the
robotics community prioritizes, or overlooks, sustainability in its formal scientific communications.

\subsection{Limitations of the Approach}

While computational analysis enables the processing of an otherwise intractable volume of 
literature (58,000 papers), it is necessary to acknowledge the inherent limitations of 
relying on an LLM as the primary annotator.

\subsubsection{The LLM is not a sustainability expert}

Because explicit sustainability motivations are typically articulated in the introduction or 
conclusion of a manuscript rather than within dense technical proofs, the cognitive reasoning 
load required for extraction is relatively moderate. However, sustainability remains a nuanced, 
multidimensional concept. While DeepSeek-V3 is highly capable, it lacks domain-specific human 
expertise. Consequently, there is an inherent risk of false negatives (overlooking subtle, 
non-standard articulations of impact) or false positives (exhibiting unwarranted optimism 
regarding a technology's sustainability potential).

In this study, the model was only ever required to extract the relevant information from the text, 
rather than to provide its "opinion" about the paper or "score" it in some way. 
However, although we grounded the model using the official text of the SDG 
targets to minimize hallucination, the model's internal weights inevitably influence its judgment. 
Unlike a human annotator utilizing a structured annotation protocol (or an expert with highly domain-specific knowledge), 
the LLM's decision-making process still remains somewhat of a "black box". To mitigate this opacity, 
we enforced a structured classification format with a strict requirement 
for textual evidence and reasoning (Point 4) to support all conclusions.

\Cref{sec:consistency} includes a detailed analysis of the consistency 
of the model's classifications across multiple runs, which provides insights 
into the repeatability and robustness of the observed trends, 
despite the inherent variability of LLM outputs.

\subsubsection{Communication vs. real-world impact}

This analysis measures scientific communication and stated intent, not actualized real-world 
impact that might happen using the developed solutions. A paper may be explicitly motivated 
by sustainability yet fail to propose a viable solution; conversely, a purely technical paper, with no 
mention of sustainability motivations, may yield breakthroughs in green energy. Therefore, this 
study quantifies the state of the field's scientific narrative, serving as a proxy for the
field's research culture and priorities, rather than its environmental footprint and implications.

\subsubsection{Statistical robustness at scale}

Despite these limitations, the primary strength of this methodology is its scale. 
While the model may misclassify individual papers, the statistical noise of these errors is 
minimized across a dataset of nearly 50,000 documents. The broad trends, pronounced gaps, 
and stagnant ratios observed are statistically robust, providing a macro-level reflection 
of the robotics community's trajectory that manual analysis could not feasibly achieve.

\Cref{sec:model_comp} includes a comparative analysis of the classification 
outputs across 4 different LLM models, which demonstrates that, despite differences in classification thresholds, 
the overall trends and patterns remain consistent. This consistency across models provides some reassurance 
that the observed trends are not solely artifacts of a single model's biases, but rather reflect 
broader patterns in the literature.

\subsubsection{Zero-shot classification limitations}

While zero-shot classification provides a highly scalable and resource-efficient method for 
processing large datasets, it is inherently constrained compared to more resource-intensive 
machine learning methodologies. Recent literature indicates that alternative approaches, 
such as few-shot prompting or model fine-tuning, frequently yield superior classification 
accuracy and better domain adaptation \citep{vajjala2025text}. 

However, compiling a 
sufficiently large, rigorously human-annotated dataset of robotics papers to serve as a 
high-quality fine-tuning training set was beyond the scope of this macro-level 
study.  By utilizing a zero-shot approach, we prioritized maximum field-wide coverage and 
methodological reproducibility, accepting the well-documented trade-offs in precision that 
accompany this technique. Future research could explore the development of 
a more accurate, domain-specific classifier through the creation of a human-annotated dataset 
and subsequent model fine-tuning, for example by taking inspiration from the ClimateQA by \citet{luccioni2020analyzing}.

It is worth noting that a few-shot prompting approach, which provides the model with a small 
number of annotated examples, could potentially enhance classification performance. 
However, this is not a trivial task, as it would require the careful selection of 
representative examples that capture the diversity of the literature, and it would 
also increase the computational cost of the analysis. This is a promising avenue for future 
research, as it could provide a more nuanced understanding of the field's sustainability 
orientation while improving the accuracy and consistency of the classifications.

\subsection{The Footprint of Our Analysis}

In alignment with the sustainability principles advocated in this work, it is important to 
transparently disclose the environmental footprint of the computational methodology itself.
Large-scale AI analysis carries substantial ecological costs.

\begin{table}[!htb]
    \centering
    \small
    \caption{Estimated resource and emissions footprint of the large-scale analysis pipeline.}
    \label{tab:analysis_footprint}
    \begin{tabular}{p{0.45\linewidth} p{0.45\linewidth}}
        \toprule
        Metric & Value \\
        \midrule
        Model Used & DeepSeek-V3 \\
        DeepSeek's Carbon Intensity Factor (CIF) & $\sim0.6$ kgCO$_2$e / kWh \citep[Table 1]{jegham2025hungry} \\
        Data Volume & 1.2 billion tokens exchanged \\
        Prompt Number & $\sim50,000$ \\
        Average Prompt Size & Input: $\sim25k$ tokens, Output: $\sim750$ tokens \\
        Estimated Energy Per Prompt & $\sim13$ Wh \citep[Table 4]{jegham2025hungry} \\
        Estimated Carbon Footprint Per Prompt & $\sim7.8$ g CO$_2$e (0.6 kgCO$_2$e/kWh $\times$ 13 Wh) \\
        Total Energy Consumption & $\sim650$ kWh \\
        Total Carbon Footprint & $\sim390$ kgCO$_2$e \\
        Total Experimental Budget & $\sim250$ \$ \\
        \bottomrule
    \end{tabular}
\end{table}

To contextualize the \textbf{650 kWh} energy expenditure:
\begin{itemize}
    \item \textbf{Yearly energy consumption per inhabitant:} This is equivalent to 29\% of the 
    average annual electricity consumption of an inhabitant in France (2,223 kWh\footnote{Data sourced from \href{https://www.data.gouv.fr/reuses/consommation-par-habitant-et-par-ville-delectricite-en-france}{data.gouv.fr}.}).
    \item \textbf{Yearly PC energy consumption:} It represents 6.5$\times$ the total annual 
    energy consumption of a standard desktop PC (assuming 100 kWh/year\footnote{Data sourced from \href{https://energy-efficient-products.ec.europa.eu/product-list/computers_en}{ec.europa.eu}.}) and 12$\times$ the consumption of a standard notebook (50 kWh/year).
\end{itemize}

To contextualize the \textbf{390 kg of CO$_2$e} emissions using common environmental equivalents:
\begin{itemize}
    \item \textbf{Driving:} Equivalent to driving a standard gasoline passenger car for 
    $\sim\!1500$ km (400gCO$_2$e/mile \footnote{Data sourced 
    from \href{https://www.epa.gov/greenvehicles/greenhouse-gas-emissions-typical-passenger-vehicle}{epa.gov}.}), 
    roughly the distance between Paris and Rome.

    \item \textbf{Flying:} Equivalent to a return-trip flight from London to Barcelona, about $\sim\!1500$ km (0.25kgCO$_2$/km \footnote{Data sourced from \href{https://skoot.eco/articles/the-true-co2-emissions-from-flying}{skoot.eco}.}).

    \item \textbf{Tree absorption:} It would require approximately 39 mature trees an entire year (10 kg/year\footnote{Data sourced from \href{https://onetreeplanted.org/blogs/stories/how-much-co2-does-tree-absorb}{onetreeplanted.org}.})
    to sequester the carbon produced by this specific experimental run.

    \item \textbf{Personal footprint:} It represents 2.6\% of the average annual carbon footprint of a US 
    inhabitant (14.5 tCO$_2$e/year\footnote{Data sourced from \href{https://ourworldindata.org/grapher/co-emissions-per-capita}{ourworldindata.org}.}), but accounts for 19\% of the targeted annual carbon footprint per capita 
    required to meet the 1.5\,$^{\circ}$C target (2 tCO$_2$e/year).
\end{itemize}

While this one-time carbon cost is substantial, we would argue that the insights generated, specifically 
identifying the \textbf{Awareness Gap} within a field that significantly influences global industry, 
constitute a necessary investment to facilitate a transition toward more sustainable research practices.

\subsection{Open-Source Code and Dataset}

To enable easier reproduction and full transparency, the complete analysis pipeline has been open-sourced, 
enabling researchers to independently evaluate their own manuscripts. The interactive \textbf{Paper Analysis Tool} 
is available on \href{https://huggingface.co/spaces/sustainable-robotics/paper-sustainability-assessment}{Hugging Face \huggingface}, 
and the underlying codebase is publicly accessible\footnote{The tool is hosted at: 
\href{https://huggingface.co/spaces/sustainable-robotics/paper-sustainability-assessment}{sustainable-robotics/paper-sustainability-assessment}}.

Furthermore, the complete dataset comprising the \textbf{about 50,000 analyzed arXiv articles}, including 
the LLM's classifications and reasoning outputs, has been released on \href{https://huggingface.co/datasets/sustainable-robotics/robotics-arxiv-sustainability-classification}{Hugging Face Datasets \huggingface}\footnote{The dataset is available at: 
\href{https://huggingface.co/datasets/sustainable-robotics/robotics-arxiv-sustainability-classification}{sustainable-robotics/robotics-arxiv-sustainability-classification}}.

\section{Analysis Results}
\label{sec:results}

The large-scale study of about 50,000 research articles within the arXiv robotics category (\texttt{cs.RO}) reveals several important 
insights into the scientific narrative and priorities of the robotics community. Our findings highlight three dominant trends:

\begin{itemize}
    \item The field of robotics is growing at an exponential rate
    \item \textbf{The Awareness Gap:} Explicit communication regarding broader impacts (social, sustainability, and ecological) 
    is marginal, with the proportion of such mentions remaining stagnant over the last decade.
    \item \textbf{The Motivation Gap:} While a vast majority of robotics research addresses topics that have the potential
    for improving sustainability, only a small fraction of authors is explicitly motivated by it.
\end{itemize}

More precisely, quantitative analysis of the dataset demonstrates that the ratio of papers mentioning social, environmental, 
or sustainability impacts has remained consistently low, at \textbf{under 6\%, 2\%, and 1.5\%}, respectively.
Additionally, the proportion of research explicitly motivated by sustainability has not exceeded \textbf{5\%}.
Finally, the engagement with global frameworks is exceptionally low, with explicit mentions of the UN SDGs appearing in \textbf{fewer than 0.1\%} of publications.

The following sections provide a detailed examination of these empirical trends.

\subsection{The Exponential Growth of Robotics Research}

\begin{figure}[!htb]
    \centering
    \includegraphics[width=0.95\linewidth]{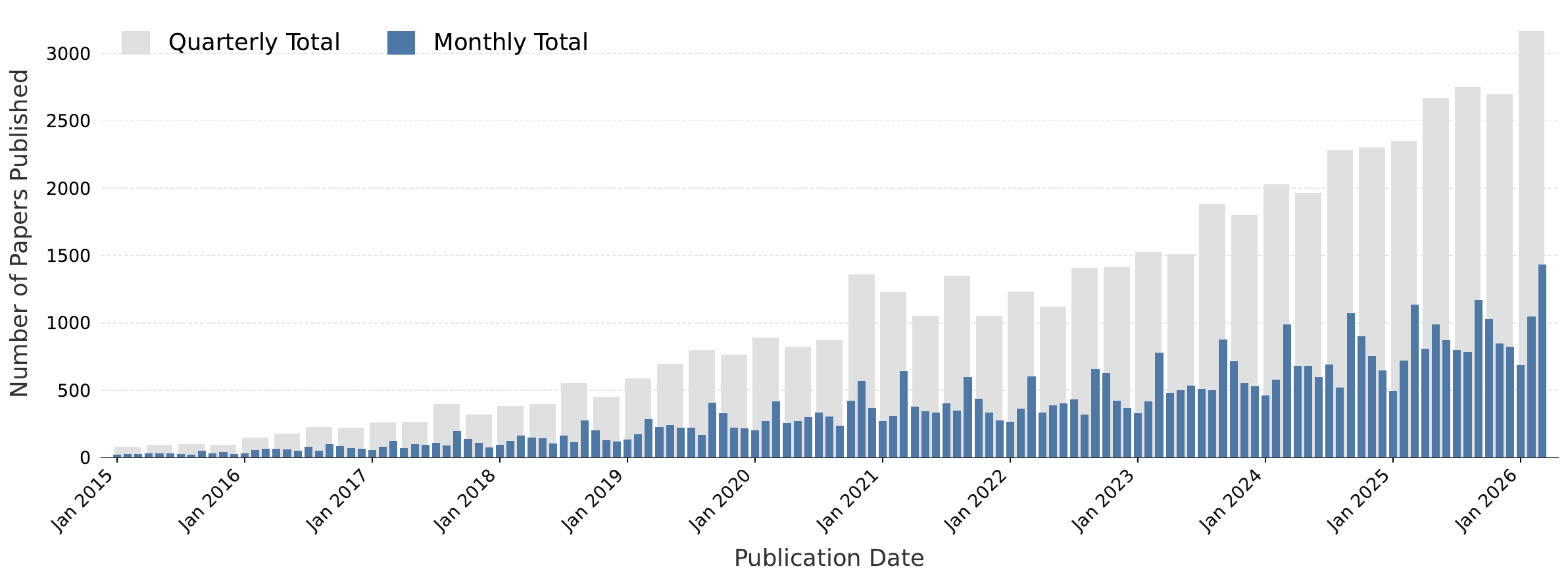}
    \caption{Total number of papers published in arXiv's \texttt{cs.RO} category over time.}
    \label{fig:total-papers-time}
\end{figure}
    
The most prominent trend identified is the massive volume of research output. As shown in \Cref{fig:total-papers-time}, the growth of the field is substantial,
increasing from 80 articles in the first quarter (Q1) of 2015 to over 3,100 articles in the first quarter (Q1) of
2026. This trajectory appears exponential, with the total publication volume doubling approximately every three years.

Periodic fluctuations can also be observed, with March and September consistently serving as peak periods for submissions, a 
trend strongly correlated with major conference deadlines, namely ICRA (September-November) and IROS (February-March).
Conversely, January represents the lowest point of annual submission cycles. Despite these seasonal variations, the 
high-level expansion of the field remains constant.

\subsection{Communication of Broader Impacts}

To assess how the research community contextualizes its work within a broader societal framework, we analyzed the 
dataset for explicit mentions of impact in three primary dimensions:

\begin{itemize}
    \item \textbf{Social Impact}
    \item \textbf{Ecological Impact}
    \item \textbf{Sustainability Topics}
\end{itemize}

\begin{figure}[!htb]
    \centering
    \includegraphics[width=0.95\linewidth]{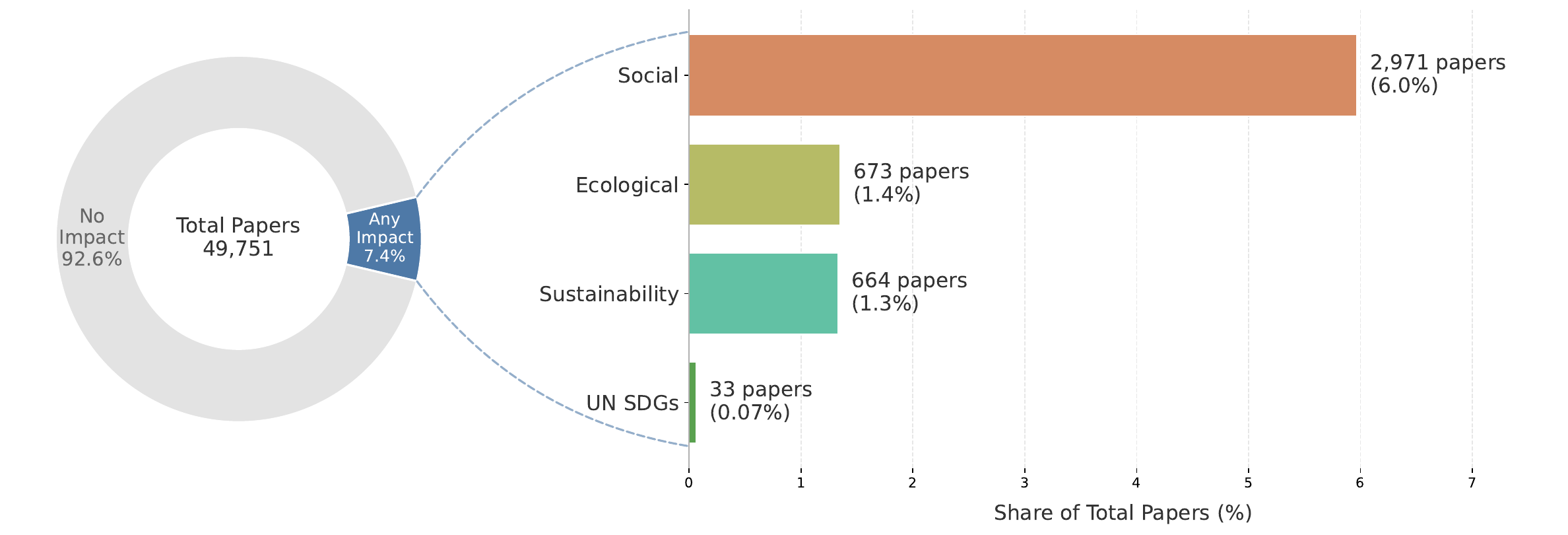}
    \caption{The ratio of papers mentioning impacts versus the one not mentioning any on the left. The right 
    graph shows the zoomed-in breakdown of impact mentions. Note: categories can overlap, one paper can mention multiple .}
    \label{fig:pie-impacts}
\end{figure}

Furthermore, we surveyed for explicit references to the \textbf{UN's Sustainable Development Goals (SDGs)} 
to evaluate the appearance of this global framework within the robotics lexicon.

It is worth noting that this analysis quantifies the \textit{presence} of explicit impact mentions in articles; it does not evaluate the depth or 
qualitative rigor of the arguments presented. Even superficial mentions or instances of \textit{"greenwashing"} are considered
as valid impact communications. 

\subsubsection{Overall Distribution of Impact Mentions}

\Cref{fig:pie-impacts} represents the analysis of the entire set of papers. The figure shows that the ratio of scientific articles 
mentioning any form of the three impacts stays under 7.5\%. More specifically, social impact mentions are the most represented, amounting to around 6\% of papers,
while sustainability and ecological impacts are mentioned in under 1.5\% of articles. Finally, the UN's SDGs are mentioned in only 0.07\% of papers (only 33 papers out of 50,000).

\subsubsection{Impact mentions in Time}

\begin{figure}[!htb]
    \centering
    \includegraphics[width=\linewidth]{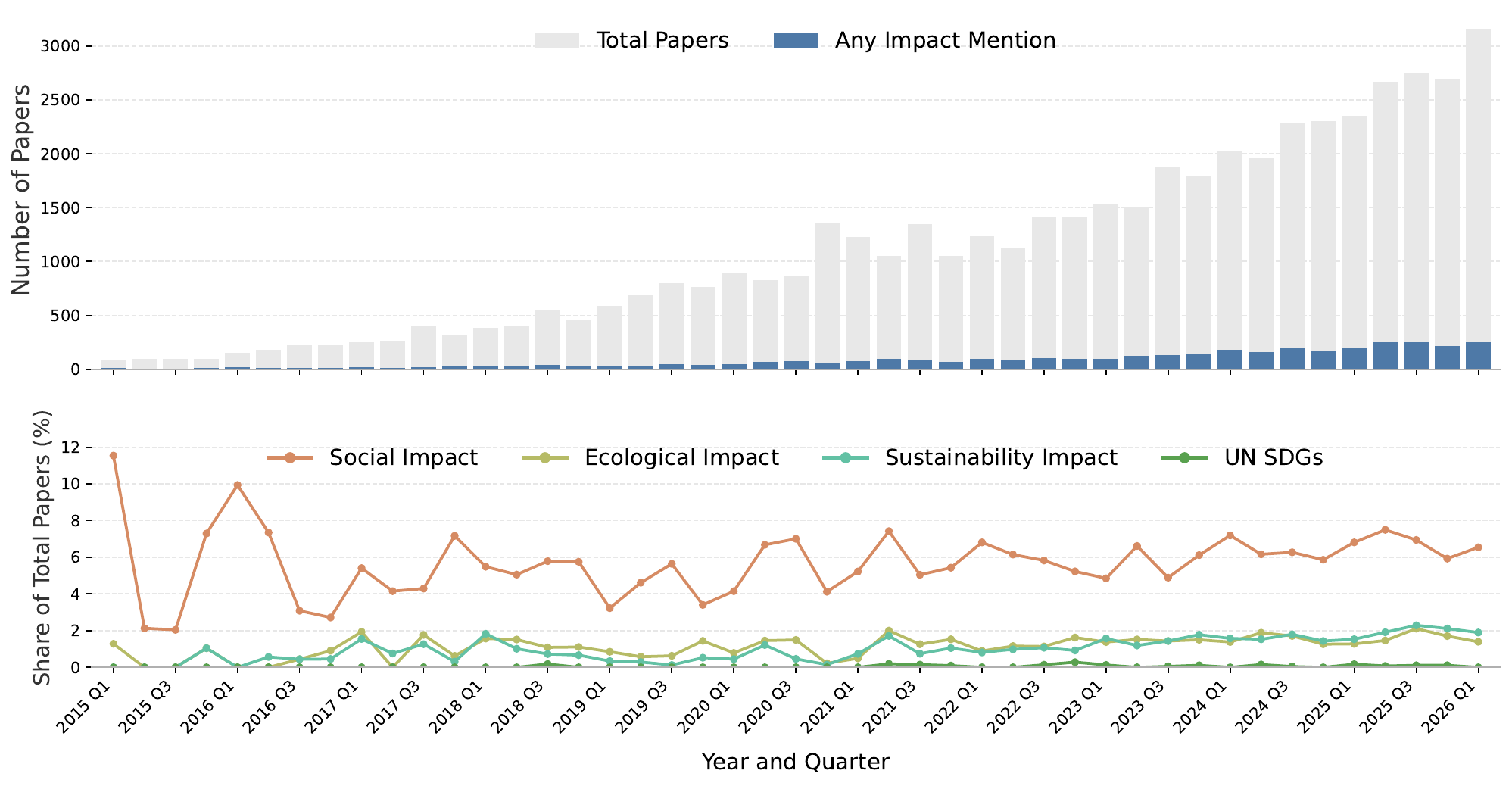}
    \caption{Quarterly impact mentions in robotics papers. The top graph compares the total number of papers with those mentioning any of the impacts. 
    The bottom graph shows the evolution of the ratio of papers mentioning the impacts and SDGs in time.}
    \label{fig:impact-quarterly}
\end{figure}

To obtain a more granular understanding of current communication practices, we analysed the evolution of the impact mentions in time. 
The data is presented in \Cref{fig:impact-quarterly} and highlights a contrast between the field's rapid expansion and its stagnant sustainability engagement.

While the absolute number of papers mentioning broader impacts has risen in time, it 
follows the trend of the field's overall growth rather than an increase in relative awareness. 
\Cref{fig:impact-quarterly} shows that the proportion of research
addressing these impacts has remained surprisingly stagnant for over ten years.

Furthermore, broader impacts seem to remain a secondary consideration in the 
literature. Mentions of social impact remain near 6\%, while environmental and sustainability impacts consistently 
fail to surpass the 2\% threshold. This is even more that case when it comes to explicit mentions of the UN SDGs in the papers, which
are nearly non-existent, appearing 
in fewer than 0.1\% of all papers. Despite serving as the international standard for sustainable development, the
SDGs are largely absent from formal robotics communications.

The empirical data presented on \Cref{fig:pie-impacts} and \Cref{fig:impact-quarterly} shows a concerning view of the field: 
a community that rarely engages with the long-term consequences of its innovation. Despite the increasing global 
urgency of sustainability, the vast majority of robotics research remains silent on its ecological and social footprint. 
In the context of this paper we refer to this phenomenon as the \textbf{Awareness Gap}, which we define as the 
persistent disconnect between the field's potential for impact and 
its explicit communication of that impact in the scientific narrative.

\subsection{Sustainability Motivation and Potential Relevance}

Moving beyond impact mentions, we evaluated the entire dataset of the articles to analyse the alignment of proposed
research topics with the topics proposed by the UN's SDGs.
To do so, we distinguish between research that is \textit{potentially applicable} to sustainability goals (\textbf{SDG Aligned}) and research that 
is \textit{explicitly driven} by those goals (\textbf{SDG Motivated}).

More precisely, using the UN SDGs as a reference framework, we classified the papers into three distinct groups:
\begin{itemize}
    \item \textbf{SDG Aligned:} Research developing technologies that facilitate an SDG (e.g., trajectory 
    optimization for industrial efficiency - SDG 9, or surgical robotics for healthcare - SDG 3).
    \item \textbf{SDG Motivated:} Aligned research that explicitly mentions sustainability as a primary 
    motivator for the work.
    \item \textbf{No Direct SDG Relevance:} Niche technical research with no identifiable link to global sustainability targets.
\end{itemize}

For example, a paper about efficient path planning for industrial robots would be classified as 
relevant for SDG 9 (SDG Aligned), but it would only be classified as SDG Motivated by SDG 9 if 
the paper explicitly mentions that the motivation of the work is to contribute to some of the SDG 
9 targets\footnote{SDG 9 targets: \url{https://sdgs.un.org/goals/goal9\#targets\_and\_indicators}} 
(e.g., “This work is motivated by the need to promote sustainable industrialization and foster innovation” - target 9.2).

As for impact mentions, we analysed the temporal evolution of the SDG Aligned and Motivated articles to evaluate the
progression of the field's awareness in time. 

\subsubsection{SDG Alignment: Distribution of SDG relevant topics}

\begin{figure}[!htb]
   \centering
    \includegraphics[width=1\linewidth]{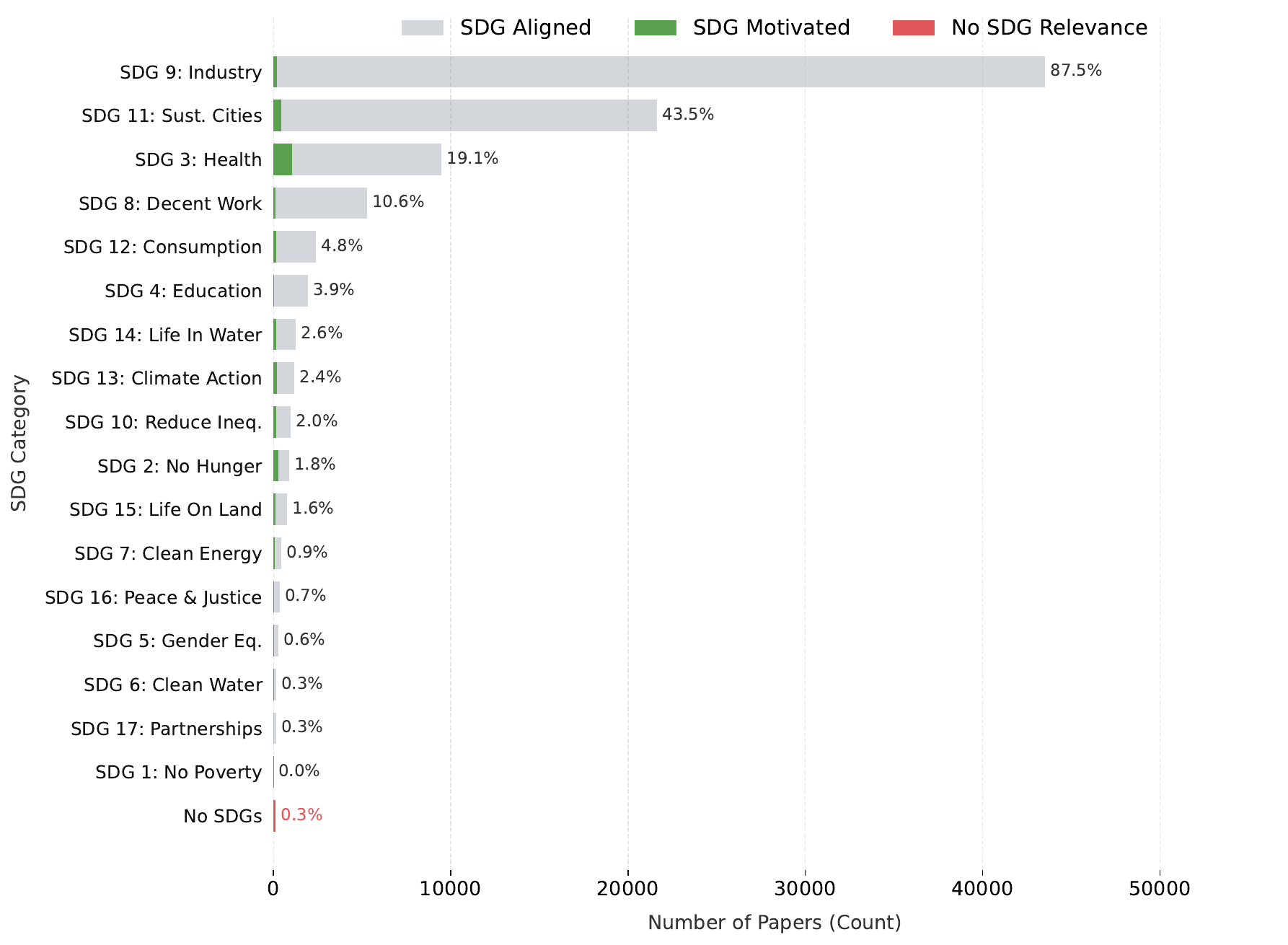}
    \caption{Number of \textbf{SDG Aligned} papers, explicitly \textbf{SDG Motivated} papers, and papers with no direct SDG relevance. 
    The percentage value shows the ratio of papers classified as \textbf{SDG Aligned} per SDG (out of all the 50,000 papers). Note: one paper can align 
    to multiple SDG topics.}
    \label{fig:sdg-explicit-indirect}
\end{figure}

\Cref{fig:sdg-explicit-indirect} presents the resulting distribution of SDG relevance (\textbf{SDG Aligned} papers), revealing a
hierarchy of covered topics in robotics research. More precisely, the figure shows the distribution 
of the papers that are classified as relevant for certain SDG topics (\textbf{SDG Aligned}). 
One article can be relevant (\textbf{SDG Aligned}) for multiple SDG topics. 

The figure shows that the vast majority of research falls under technological advances for
Industrial Innovation (SDG 9) at 88\%, followed by Sustainable Cities and Communities (SDG 11) at 44\%, 
Decent Work and Economic Growth (SDG 8) at 10\%, and Responsible Consumption and Production (SDG 12) at 5\%. 
These areas are driven by industrial applications, process optimization, and manufacturing efficiency.
The third-largest category is Good Health and Well-being (SDG 3) with approximately 19\% of papers, propelled
by the sub-field of medical and assistive robotics.

In contrast, environmental domains such as Climate Action (SDG 13), 
Life on Land (SDG 15), Life Below Water (SDG 14), and Clean Energy (SDG 7) are significantly marginalized,
with each accounting for less than 3\% of the corpus. Furthermore, social dimensions, including No Poverty (SDG 1), Peace, Justice and Strong 
Institutions (SDG 16), Gender Equality (SDG 5), and Reduced Inequalities (SDG 10), are similarly 
underrepresented (less than 2\% each).

Only 0.3\% of papers were classified as having no direct SDG relevance. These results
confirm that the field possesses the inherent potential to address global challenges, but it also demonstrates that the UN 
SDGs serve as a broad framework capable of encompassing nearly any technical advancement.

It is important to recognize that these results are sensitive to the classification methodology. While a more 
conservative LLM prompting strategy might yield a lower overall proportion of SDG relevance, such a shift would
likely impact the SDG Aligned and SDG Motivated values equally, leaving the substantial gap between technical 
potential and explicit intent unchanged.

\Cref{fig:sdg-explicit-indirect} further hints that the overall volume of research explicitly motivated 
by SDG topics (\textbf{SDG Motivated}) is severely disproportionate to the volume of research treating 
topics relevant to certain SDGs (\textbf{SDG Aligned}). \Cref{fig:sdg-explicit-indirect-relative} and the subsequent discussion 
examine this relationship in greater detail.

See the appendix (\Cref{sec:appendix_landscape}) for a visual representation of the semantic landscape of SDG Aligned papers in 
robotics literature.

\subsubsection{SDG Motivation: Distribution of SDG motivated articles}

\begin{figure}[!htb]
    \centering
    \includegraphics[width=0.95\linewidth]{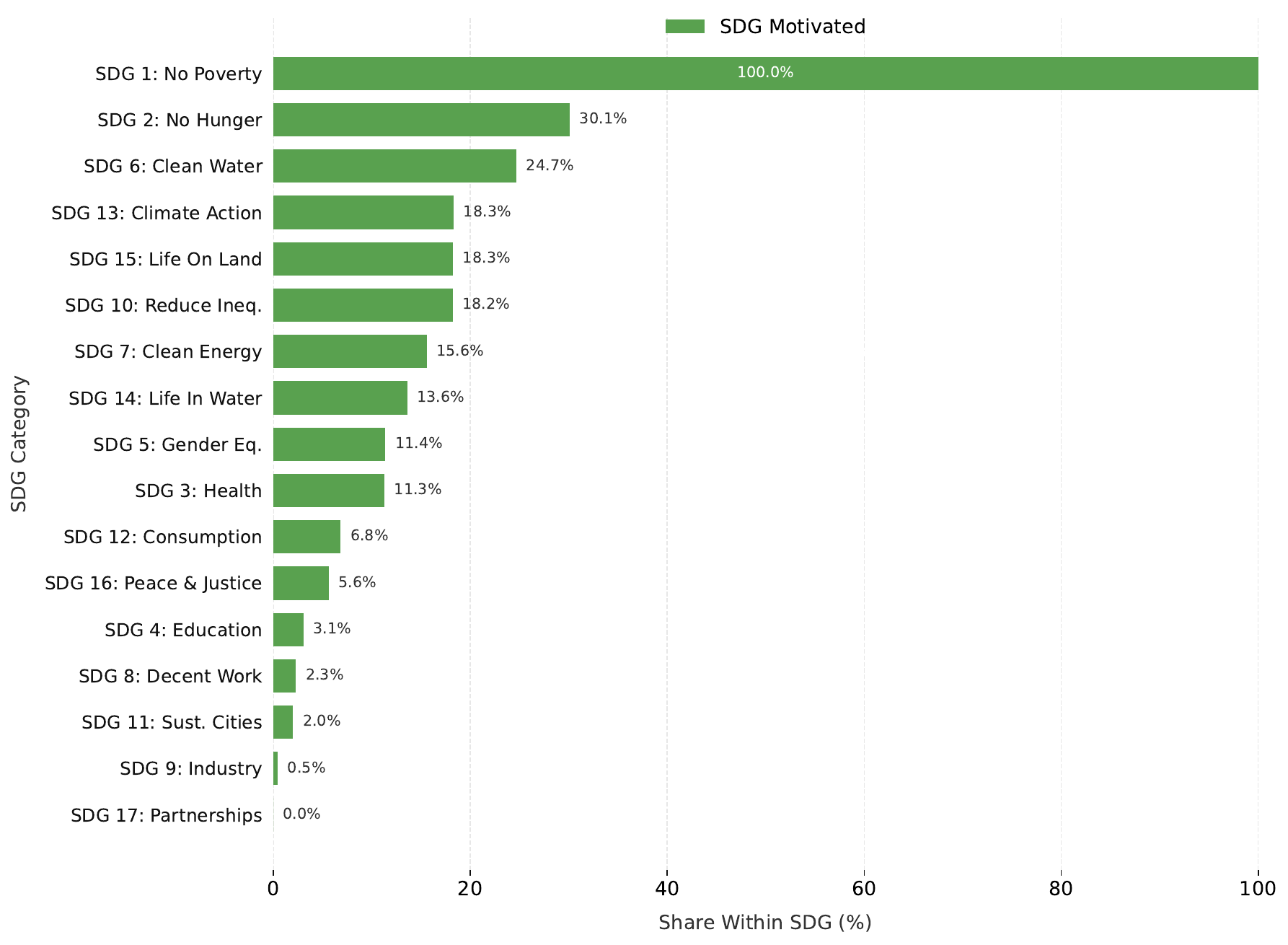}
    \caption{Share of explicitly \textbf{SDG Motivated} papers relative to the total \textbf{SDG Aligned} papers per SDG topic.}
    \label{fig:sdg-explicit-indirect-relative}
\end{figure}

Having the list of the SDG Aligned and SDG Motivated articles for each SDG topic
allows us to quantify the sustainability motivation of research articles per individual SDG topics.
The results in \Cref{fig:sdg-explicit-indirect-relative} demonstrate a significant 
"Motivation Disparity" across domains:

Human well-being sectors such as Health (SDG 3), Zero Hunger (SDG 2), Clean Energy (SDG 7), 
and Clean Water and Sanitation (SDG 6) exhibit relatively high rates of explicit motivation (typically well 
above 10\%)\footnote{No Poverty (SDG 1) shows a 100\% motivation rate; however, due to a sample size of only one paper, this result constitutes a statistical outlier.}. 
Notably, SDG 2 (Zero Hunger) records the highest relative proportion of explicit motivation (30\%),
reflecting a prevalent focus on technological solutions in agriculture.
Furthermore, environmental SDGs: Climate Action (SDG 13), Life on Land (SDG 15), and Life Below Water (SDG 14), 
also demonstrate strong explicit motivations (>10\%), likely due to the increasing global visibility of 
environmental degradation. 

Social SDGs: Reduced Inequalities (SDG 10), Gender Equality (SDG 5), and Peace, Justice and Strong 
Institutions (SDG 16) have relatively moderate motivation proportions (>5\%). Followed by Responsible Consumption and Production (SDG 12) and Decent Work and 
Economic Growth (SDG 8), with approximately 2.5-7\% of papers are explicitly motivated by these goals.
Finally, Quality Education (SDG 4) sits at a relatively low 3\% as well. These results are somewhat surprisingly 
low, given the critical importance of all these considerations and their direct impact on human lives.

Somewhat expectedly, in technical/industrial domains such as Industrial Innovation (SDG 9) and Sustainable Cities and 
Communities (SDG 11), the proportion of explicitly motivated papers is very low, well under 
2\%. Research in these areas remains almost exclusively driven by technical challenges and industrial 
performance metrics rather than sustainability outcomes. 

The results suggest that the robotics community has a high potential to contribute to sustainability goals, 
but the motivation to do so is unevenly distributed across different domains. The domains 
with the direct potential sustainability impact (e.g., health, environment) are also the ones with the 
highest proportion of \textbf{SDG Motivated} papers, while the domains with less obvious and somewhat indirect potential sustainability 
impact (e.g., industrial and technical domains) are also the ones with the lowest proportion of SDG motivation.

This \textit{"Motivation Disparity"} highlights the need for greater awareness and incentives to encourage researchers in all domains 
to consider the sustainability implications of their work and to explicitly align their motivations 
with global sustainability goals.

\subsubsection{Temporal Evolution of SDG Motivation}

\begin{figure}[!htb]
    \centering
    \includegraphics[width=0.95\linewidth]{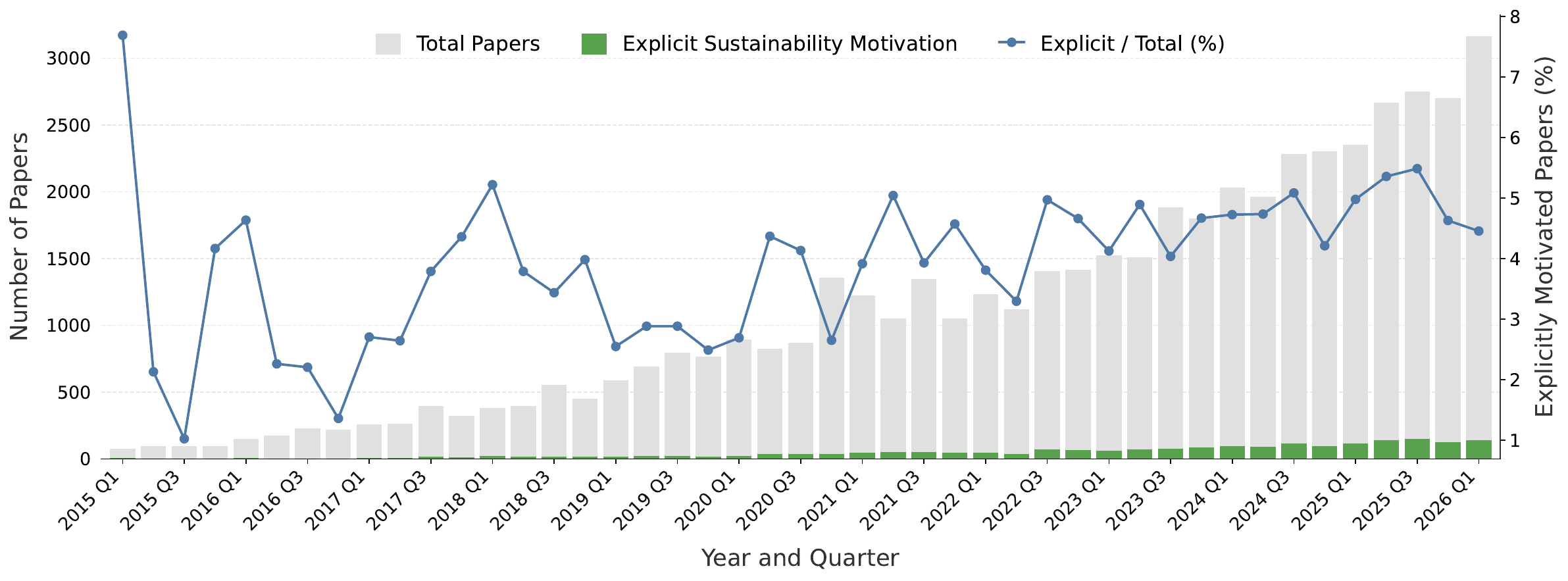}
    \caption{Comparison of total papers versus those motivated by sustainability, with percentage trend line.}
    \label{fig:explicit-vs-total}
\end{figure}

Finally, \Cref{fig:explicit-vs-total} illustrates the temporal trend of sustainability-motivated research,
(\textbf{SDG Motivated}) papers aggregated across all SDGs, relative to total publication output. 

The results show that the absolute number of motivated papers is increasing, however this growth is proportional to the overall expansion of the 
field, resulting in a persistent, flat percentage trend line. The ratio of explicitly motivated papers hovers 
at \textbf{under 5\% on average}, confirming that the vast majority (\textbf{above 95\%}) of robotics research 
is published without any explicit sustainability framing. 
 In the context of this paper, we refer to this phenomenon as the \textbf{Motivation Gap}, 
which we define as the persistent disconnect between the field's potential for sustainability impact 
and its explicit motivation by sustainability goals in the scientific narrative.

Given the urgency of the recent global challenges~\citep{kitzmann2025planetary,Richardson2023} and the enormous 
potential for robotics to address them~\citep{haidegger2023robotics,guenat2022meeting}, 
the current trajectory is insufficient. The research community must move beyond "accidental sustainability" 
and intentionally integrate these global goals into the foundational motivation of its scientific inquiries.
\section{Discussion: Closing the Sustainability Gap}
\label{sec:conclusion}

The empirical evidence presented in this study indicates that robotics research is currently not \textit{"thinking much"} about sustainability.
Coupled with the new AI capabilities, the field is experiencing an unprecedented time of technical advancement, however, 
this growth remains largely decoupled from the most pressing global challenges of our time~\citep{meadows2018limits}.

Our analysis of 50,000 articles has quantified a persistent gap in the field's sustainability awareness.
The robotics community is developing technologies with the inherent potential to drive a global sustainable transition, yet 
this relevance is currently likely accidental, rather than intentional design.
\textbf{More than 95\% of research articles do not integrate sustainability} when motivating their research (\textbf{Motivation Gap}),
and the proportion of researchers explicitly mentioning the environmental, social and sustainability-related impacts 
has remained stagnant for over a decade, typically accounting for \textbf{well under 2\% of all publications}  (\textbf{Awareness Gap}).

\textbf{If the field is to meaningfully contribute to global sustainability targets, 
incidental alignment is no longer sufficient.}
We must transition toward a paradigm of intentional impact.

\subsection{A Roadmap for Change}

Scientists are increasingly recognizing the need to align research with global sustainability 
targets, as evidenced by the growing number of public calls to action across scientific community.
Some of the most well known being \textit{“World Scientists' Warning to Humanity”} signed by more than 15,000 
scientists~\citep{scientists2017world} and \textit{"World Scientists' Warning of a Climate Emergency"}
signed by more than 11,000 scientists~\citep{ripple2020world}.

Transitioning the any academic culture, and especially technology-driven ones like robotics,
requires a combination of individual researcher actions 
and institutional reform. Below, we propose a set of concrete steps to bridge the identified \textbf{Awareness} 
and \textbf{Motivation Gaps} within the community. It is important to note that these recommendations 
are not in any way exhaustive, they are intended to be a starting point for discussion and action. 

\subsubsection{Formalizing Impact and SDG Statements}

The most immediate method to address the Awareness Gap is to formalize the integration of impact assessments 
into the scientific narrative. We advise authors to include dedicated sections, in their future publications, addressing:

\begin{itemize}
    \item \textbf{Broader impacts:} Explicitly mentioning the social, ecological, and sustainability impacts of the proposed technology.
    \item \textbf{Formal SDG alignment:} Identifying specific UN Sustainable Development Goals supported by the research.
    \item \textbf{Intentional framing:} Proactively aligning research motivations with sustainability targets at 
    the project's inception, rather than as a post-hoc justification.
\end{itemize}

If researchers are unsure which SDGs their research aligns with, they may utilize the interactive \textbf{Paper Analysis Tool} 
developed in this study to classify their work. 
The tool is available on \href{https://huggingface.co/spaces/sustainable-robotics/paper-sustainability-assessment}{Hugging Face \huggingface}\footnote{The tool is hosted at: 
\href{https://huggingface.co/spaces/sustainable-robotics/paper-sustainability-assessment}{sustainable-robotics/paper-sustainability-assessment}}.

\subsubsection{Open-Sourcing Research}

Open-source contributions represent a highly actionable mechanism for supporting a sustainable future.

One of the primary concerns regarding the future of robotics is the 
amplification of global inequalities \citep{haidegger2023robotics,guenat2022meeting}. Open-sourcing 
research democratizes access, ensuring advanced technologies are available for global development rather than restricted to privileged institutions.

Open-Sourcing improves scientific transparency. Publicly available code and datasets allow for rigorous ethical 
and sustainability auditing before technologies reach industrial scale.

Furthermore, partnerships and collaborative knowledge sharing is central to sustainability, 
as highlighted in SDG 17 (Partnerships for the Goals). Notably, our study found (\Cref{fig:sdg-explicit-indirect}) 
that SDG 17 is the only goal with zero explicitly motivated papers; open-source initiatives provide the necessary infrastructure to fill this gap.

Last but not lest, by reducing the need to \textit{"reinvent the wheel"}, open-source practices 
minimize the redundant consumption of energy, compute resources, and human capital, thereby accelerating collective progress~\citep{Bertram2020}.

\subsubsection{Diversifying Research Towards Underserved SDGs}

As demonstrated in \Cref{sec:results}, robotics research is currently heavily concentrated within industrial 
innovation (SDG 9) and urban infrastructure (SDG 11). While recent literature indicates that robotics possesses the 
potential to contribute across a wide range of SDGs \citep{haidegger2023robotics}, many of these domains remain
critically underexplored. This represents a strategic opportunity. As highlighted in \Cref{fig:sdg-explicit-indirect}, 
we identify a need for increased community focus on:

\begin{itemize}
    \item \textbf{SDGs 13, 14, and 15:} Climate action, marine conservation, and terrestrial ecosystem protection.
    \item \textbf{SDGs 5 and 10:} Reducing gender and structural inequalities.
    \item \textbf{SDG 2:} Enhancing food security through sustainable, precision agricultural robotics.
\end{itemize}

Furthermore, exploration is highly encouraged in almost any domain outside the heavily saturated areas of SDG 9 and SDG 11.


\subsubsection{Encouraging Community Discourse}

A contributing factor to the persistence of the Awareness and Motivation Gaps is the limited presence of these topics 
in standard professional discourse. To incentivise a cultural shift and encourage broader consideration of these factors, 
sustainability must be normalized within laboratory environments, academic conferences, and formal publications.

One of the most direct ways to engage is participating in workshops, panels, seminars, and conferences. 
Such events help to raise the community awareness and signals that sustainability is a community priority.
Furthermore, when possible, we encourage authors to actively discuss the alignment of research with global challenges, 
and explicitly mention the importance of sustainability in publications, presentations, and public communications.

Engaging in these discussions establishes a supportive community for researchers prioritizing these critical issues. 
Normalizing the foundational \textit{intent} behind our research is the first step toward a more intentional field.

\subsubsection{Institutional and Structural Reforms}

To achieve systemic and long-term transformation, individual agency must be reinforced by 
institutional support. We advocate for major flagship venues, such as \textbf{ICRA} and \textbf{IROS}, to adopt 
more explicit sustainability-oriented measures:
\begin{itemize}
    \item \textbf{Commitment to transparency:} Formalizing the disclosure of sustainability 
    engagement and the environmental footprints of conferences and large-scale experiments.
    \item \textbf{Implementing guiding policies:} Encouraging, or mandating, authors to evaluate 
    the broader sustainability implications of their research.
\end{itemize}

However, more ambitious structural reforms should also be considered:
\begin{itemize}
    \item Adding \textbf{sustainability as a formal review criterion}.
    \item Establishing additional \textbf{sustainability-focused tracks} and workshops.
    \item \textbf{Mandating impact statements} or sustainability assessments for all submissions.
\end{itemize}

When flagship conferences prioritize these values, the broader community inevitably follows. 

Strong positive precedents have been set by \href{https://humanrobotinteraction.org/2025/robots-for-a-sustainable-world/}{IEEE HRI 2025}, 
which featured a special theme on Robots for a Sustainable World and a dedicated Sustainability Chair, 
and \href{https://humanrobotinteraction.org/2026/}{IEEE HRI 2026}, which introduced sustainability recognition for papers. 
These represent significant milestones in the right direction.

\subsection{Concluding Remarks}

With seven of the nine planetary boundaries already surpassed~\citep{kitzmann2025planetary}, 
the paradigm of \textit{"blind progress"}, the assumption 
that technology should be developed simply because it 
\textit{can} be, is no longer scientifically or ethically tenable.

Research tells us that robotics has the potential to be a powerful force for good, 
but only if we intentionally steer it in that direction~\citep{haidegger2023robotics,guenat2022meeting}.

\textbf{Let us choose to be part of the solution, not part of the problem.}
\newpage
\bibliographystyle{hfstyle/plainnat}

\bibliography{sustainability}

\newpage
\appendix
\section{Appendix: Prompt Details and Examples}
\label{sec:appendix}

To ensure consistent zero-shot classification across the entire dataset, the DeepSeek-V3 Large 
Language Model (LLM) was provided with the full text of the official UN SDG target\footnote{The source reference used in the prompts is available at: 
\href{https://github.com/huggingface/arxiv-robotics-sustainability-classification/blob/main/data/prompts/un_sdgs.txt}{Download UN SDG Targets Text}}
. The classification pipeline used a two-part prompting strategy: a 
comprehensive \textit{System Prompt} to establish the main classification rules,  
provide the full classification taxonomy, and define the 
decision-making framework. The system prompt is then followed by an \textit{Individual Paper Prompt} applied 
to the full text of each manuscript.

\paragraph{System Prompt Architecture}
The system prompt was engineered to establish the model as a classification expert 
and prepare it for sustainability analysis. The prompt provides the model with the taxonomy of labels
and a detailed description of the classification criteria.  
In addition to the UN's SDG text we provided the model with the International Federation of Robotics (IFR) 
proposals on how robots can help achieve the SDGs\footnote{The source reference used in the prompts is available at: 
\href{https://github.com/huggingface/arxiv-robotics-sustainability-classification/blob/main/data/prompts/ifc_sdg_proposals.txt}{Download IFR SDG Proposals Text}}.
The model is also provided with a rough structure of the expected response. 

\begin{verbatim}
My aim is to assess and quantify how often robotics research is explicitly motivated 
by sustainable development, to raise awareness about the prioritization of sustainability 
in the field. You are an expert in analyzing academic papers for their impact on 
the UN Sustainable Development Goals (SDGs) and related environmental or social impacts.
Your task is to analyze the provided paper and identify which SDGs it
most directly supports, based on its content.

Use the full list of 17 SDGs and targets:
{sdg_text}

Also use the following text as a reference for the SDGs and targets. 
It shows the proposals of the International Federation of Robotics (IFR) 
on how robots can help achieve the SDGs:
{ifr_text}


Respond in the format:

Point 0. provide the type of the paper (survey, experimental, theoretical, report or other - 
or a combination of these).
If you are not sure, just say "other".
Point 1. SDGs and targets the paper is explicitly motivated by or aims to address (i.e., the 
         problem or impact the authors are directly targeting).
Point 2. provide a list of SDGs and targets SDGs and targets relevant to the technologies or 
         methods developed in the paper, even if not mentioned or motivated by sustainability. 
         This list should include SDGs the list in point 1.
Point 3. check if the terms "sustainability", "ecological impact", "social impact" or their 
         derivatives are mentioned in the text, and provide a yes/no answer for each. 
         Also check if the authors mention the UN's 17 sustainable development goals explicitly.
Point 4. **IFR Proposals:** Does the paper results/technology coincide with the International 
        Federation of Robotics (IFR) proposals for supporting SDGs? Provide a list of the the 
        SDGs the paper supports according to the IFR proposals, and a list of matching IFR 
        use cases (quote or paraphrase from IFR proposals). If the paper does not match any 
        IFR proposal, provide an empty list.
Point 5. provide a reasoning for the choices made in points 0-3, with quotes from the paper if 
         possible, make it concise and to the point.
\end{verbatim}

\paragraph{Paper-Level Extraction Logic}
Each extracted manuscript was processed using a structured prompt designed to enforce a standardized, 
reproducible classification schema. This prompt specifically requires the model to 
distinguish between \textbf{SDG Aligned} papers, the papers working on a technology that
could theoretically be used for some of the SDGs and \textbf{SDG Motivated} papers, where authors explicitly
motivate their research by certain SDG topic. Furthermore, by forcing the model to provide direct quotes and a 
concise reasoning statement, we ensure that the classification remains anchored to the 
source text, allowing for post-hoc human validation.

In addition to the SDG classification, the model was prompted to determine if the paper's topic aligns with one of the 
International Federation of Robotics (IFR) proposals for supporting SDGs, providing a direct link between 
the paper's content and practical applications in sustainable robotics. 
Although this information is not directly the main objective of the analysis, it provides an additional information
for the model to determine the sustainability relevance of the paper, and it also allows us to explore the 
potential of robotics research to contribute to the SDGs, even if not explicitly motivated by sustainability.

\begin{verbatim}
Title: {paper_title}
Full paper text:
"""
{paper_text}
"""

Do not be verbose, keep it concise.
Do not be overly optimistic or pessimistic, just state the facts.
If not enough information is available, say so and provide with
empty lists or "unknown" where appropriate.

Respond in the format:
---------------------------
0. Paper type: [survey, experimental, theoretical, report, other]
1. SDGs and targets the paper is explicitly motivated by or aims
   to address (i.e., the problem or impact the authors are directly targeting)
   only if they are motivated by sustainability not the technology itself:
   - SDGs: [SDG X, SDG Y, ...]
   - Targets: [[X.Y, X.Z, ...], [X.Y, ...], ...]
   - Quote(s) from the motivation/introduction.
2. SDGs and targets relevant to the technologies or methods developed in
   the paper, even if not motivated by sustainability but mentioned in the text:
   - SDGs: [SDG X, SDG Y, ...]
   - Targets: [[X.Y, X.Z, ...], [X.Y, ...], ...]
   - Brief justification for each.
3. Authors mention in the text:
   - UN SDGs: yes/no
   - Sustainability impact: yes/no
   - Ecological impact: yes/no
   - Social impact: yes/no
4. **IFR Proposals:** Does the paper results/technology coincide with the 
International Federation of Robotics (IFR) proposals for supporting SDGs?  
   - IFR-aligned SDGs/targets: [SDG X, SDG Y, ...]  
   - Matching IFR use cases (quote or paraphrase from IFR proposals):  
     - [E.g., “Robots used in the development & testing of drugs” or 
     “Inspection robots enable leak detection in pipes”]  
   - Brief justification/explanation.
5. Reasoning: "A concise summary of why these SDGs and alignments
   were chosen, quoting the paper where possible."
------------------------
\end{verbatim}

To demonstrate the model's ability to differentiate between technical potential and explicit intent, 
\Cref{tab:methodology_examples} compares two representative manuscripts. \textbf{Example A} illustrates 
a common "Motivation Gap", where high-tech research is aligned with innovation (SDG 9) but lacks any 
sustainability framing. \textbf{Example B} represents the small subset of research that is explicitly 
sustainability-driven.

\begin{table}[!htb]
    \centering
    \small
    \caption{Comparison of model behavior on a purely technical paper versus a sustainability-driven paper.}
    \label{tab:methodology_examples}
    \begin{tabular}{p{0.16\linewidth} p{0.38\linewidth} p{0.38\linewidth}}
        \toprule
        Feature & \textbf{Example A: Purely Technical} & \textbf{Example B: Sustainability Driven} \\
        \midrule
        Title &	Factorization of Rational Motions: A Survey with Examples and Applications & Adaptive Sensor Placement Inspired by Bee Foraging: Towards Efficient Environment Monitoring \\
        Paper Link & \href{http://arxiv.org/abs/1501.06862}{arXiv:1501.06862} & \href{https://arxiv.org/abs/2411.15159v1}{arXiv:2411.15159} \\
        Motivated by SDG & No & Yes (SDG 15, SDG 13)  \\
        Aligned to SDGs & SDG 9 (Innovation) & SDG 15 (Ecosystems), SDG 13 (Climate),SDG 9 (Innovation) \\
        Impacts Mentioned & None & ``Sustainability'', ``Ecological Impact'' \\
        Model Reasoning & The paper is a survey on the factorization of rational motions for mechanism 
synthesis, with examples including Bennett linkages and open chains. Its motivation is purely 
academic and technical within kinematics, with no mention of sustainability, social, or 
ecological impacts. The developed mathematical methods are relevant to advancing 
engineering and innovation (SDG 9.5), but this is not an explicit aim. There is no 
direct alignment with the specific robotic applications listed in the IFR proposals. & The paper is explicitly motivated by addressing climate change and 
environmental degradation through sustainable robotics for forest monitoring (SDG 13, 15). 
The developed hybrid ABC-Levy algorithm for optimizing sensor placement in swarms is a 
technological innovation relevant to SDG 9. The authors mention 'sustainable robotics' 
and 'ecological' impact but do not cite the UN SDGs. The technology aligns with IFR's SDG 
15 proposals for using robots to protect ecosystems via environmental monitoring, though 
it focuses on data gathering rather than active soil or chemical management.\\
        \bottomrule
    \end{tabular}
\end{table}

\paragraph{Example A: Received response}
\begin{verbatim}
---------------------------
0. Paper type: survey

1. SDGs and targets the paper is **explicitly motivated by or aims to address** (i.e., 
the problem or impact the authors are directly targeting) only if they are motivated by 
sustainability not the technology itself:
    - SDGs: []
    - Targets: []
    - Quote(s) from the motivation/introduction: The paper's abstract and introduction 
    focus on \"factorization theory for rational motions\" and its applications in \"theoretical 
    and applied mechanism science,\" \"mechanism synthesis,\" and \"linkage synthesis.\" There 
    is no explicit mention of sustainability, environmental, or social problems as motivation.

2. SDGs and targets **relevant to the technologies or methods developed** in the paper, even if 
not motivated by sustainability but mentioned in the text:
    - SDGs: [SDG 9]
    - Targets: [[9.5]]
    - Brief justification for each: The paper develops mathematical and algorithmic tools for 
    mechanism synthesis (e.g., for Bennett linkages, 6R linkages). This contributes to 
    \"technological capabilities\" and \"innovation\" in mechanical engineering and robotics, 
    aligning with SDG 9 (Industry, Innovation and Infrastructure), specifically target 9.5 
    (\"Enhance scientific research, upgrade the technological capabilities of industrial 
    sectors... encouraging innovation\").

3. Authors mention in the text:
    - UN SDGs: no
    - Sustainability impact: no
    - Ecological impact: no
    - Social impact: no

4. **IFR Proposals:** Does the paper results/technology coincide with the International 
Federation of Robotics (IFR) proposals for supporting SDGs?  
    - IFR-aligned SDGs/targets: []  
    - Matching IFR use cases (quote or paraphrase from IFR proposals): []  
    - Brief justification/explanation: The paper is a theoretical/survey paper on motion 
    factorization for mechanism design. While the resulting linkages could theoretically be 
    components in robotic systems, the paper itself does not develop or propose specific 
    robotic applications that match the IFR's use cases (e.g., agricultural robots, 
    inspection robots, etc.). The work is foundational mathematics for kinematics.

5. Reasoning: \"The paper is a survey on the factorization of rational motions for mechanism 
synthesis, with examples including Bennett linkages and open chains. Its motivation is purely 
academic and technical within kinematics, with no mention of sustainability, social, or 
ecological impacts. The developed mathematical methods are relevant to advancing 
engineering and innovation (SDG 9.5), but this is not an explicit aim. There is no 
direct alignment with the specific robotic applications listed in the IFR proposals.\"
---------------------------
\end{verbatim}

\paragraph{Example B: Received response}
\begin{verbatim}
---------------------------
0. Paper type: experimental

1. SDGs and targets the paper is **explicitly motivated by or aims to address**:
    - SDGs: [SDG 13, SDG 15]
    - Targets: [[13.1, 13.3], [15.1, 15.5]]
    - Quote(s) from the motivation/introduction:
    - \"The challenges of climate change and environmental degradation call for 
    innovative solutions that can support sustainable practices across various fields, including robotics.\"
    - \"Forest canopies, in particular, are critical for biodiversity, water cycles, 
    and carbon storage. Monitoring these areas helps scientists understand ecosystem 
    dynamics, assess forest health, and identify threats...\"
    - \"This work contributes to sustainable robotics and lays the groundwork for 
    conservation, resource management, and climate monitoring applications.\"

2. SDGs and targets **relevant to the technologies or methods developed** in the paper, 
even if not motivated by sustainability but mentioned in the text:
    - SDGs: [SDG 13, SDG 15, SDG 9]
    - Targets: [[13.1, 13.3], [15.1, 15.5], [9.5]]
    - Brief justification for each:
    - SDG 13 (Climate Action): The paper's core application is environmental monitoring 
    for climate change understanding and forest health assessment.
    - SDG 15 (Life on Land): The technology is designed for monitoring terrestrial 
    ecosystems (forest canopies) to protect biodiversity and assess forest health.
    - SDG 9 (Industry, Innovation and Infrastructure): The paper develops a novel 
    hybrid algorithm (innovation) for optimizing robotic sensor placement, 
    enhancing technological capabilities for environmental monitoring.

3. Authors mention in the text:
    - UN SDGs: no
    - Sustainability impact: yes
    - Ecological impact: yes
    - Social impact: no

4. **IFR Proposals:** Does the paper results/technology coincide with the 
International Federation of Robotics (IFR) proposals for supporting SDGs?
    - IFR-aligned SDGs/targets: [SDG 15]
    - Matching IFR use cases (quote or paraphrase from IFR proposals):
    - \"Robots help to limit the need for chemical control and fertilizers with a 
    high degree of predictability and precision, protecting the ecosystem and reducing the impact on soil dwellers.\"
    - \"Mobile robots preventing soil erosion...\"
    - Brief justification/explanation: The paper's focus on using a robotic swarm 
    for efficient, minimally invasive forest canopy monitoring aligns with the IFR's 
    description of robots protecting terrestrial ecosystems through precise environmental 
    monitoring and data collection, though the paper's application is more about 
    monitoring than direct intervention like limiting chemicals.

5. Reasoning: \"The paper is explicitly motivated by addressing climate change and 
environmental degradation through sustainable robotics for forest monitoring (SDG 13, 15). 
The developed hybrid ABC-Levy algorithm for optimizing sensor placement in swarms is a 
technological innovation relevant to SDG 9. The authors mention 'sustainable robotics' 
and 'ecological' impact but do not cite the UN SDGs. The technology aligns with IFR's SDG 
15 proposals for using robots to protect ecosystems via environmental monitoring, though 
it focuses on data gathering rather than active soil or chemical management.\"
---------------------------
\end{verbatim}

\newpage
\section{Appendix: Landscape of Sustainability in robotics research}\label{sec:appendix_landscape}
\begin{figure*}[!htb] 
    \centering
    
    \begin{minipage}[c]{0.64\textwidth}
        \begin{subfigure}{\textwidth}
            \includegraphics[width=\textwidth]{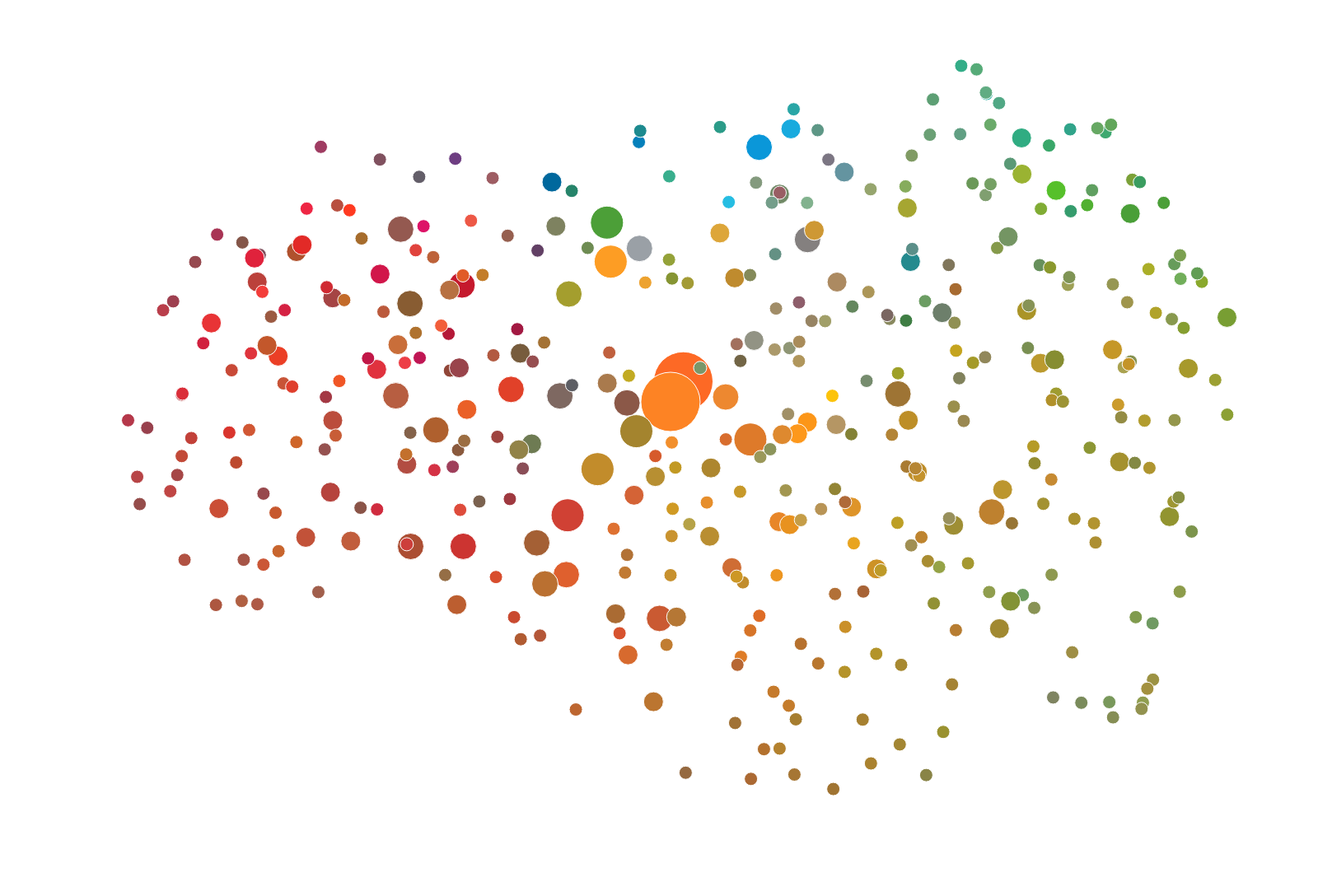}
            \caption{\textbf{All Papers (48,163, 100\%)}}
        \end{subfigure}
    \end{minipage}
    \hfill
    \begin{minipage}[c]{0.32\textwidth}
        \begin{subfigure}{\textwidth}
            \includegraphics[width=\textwidth]{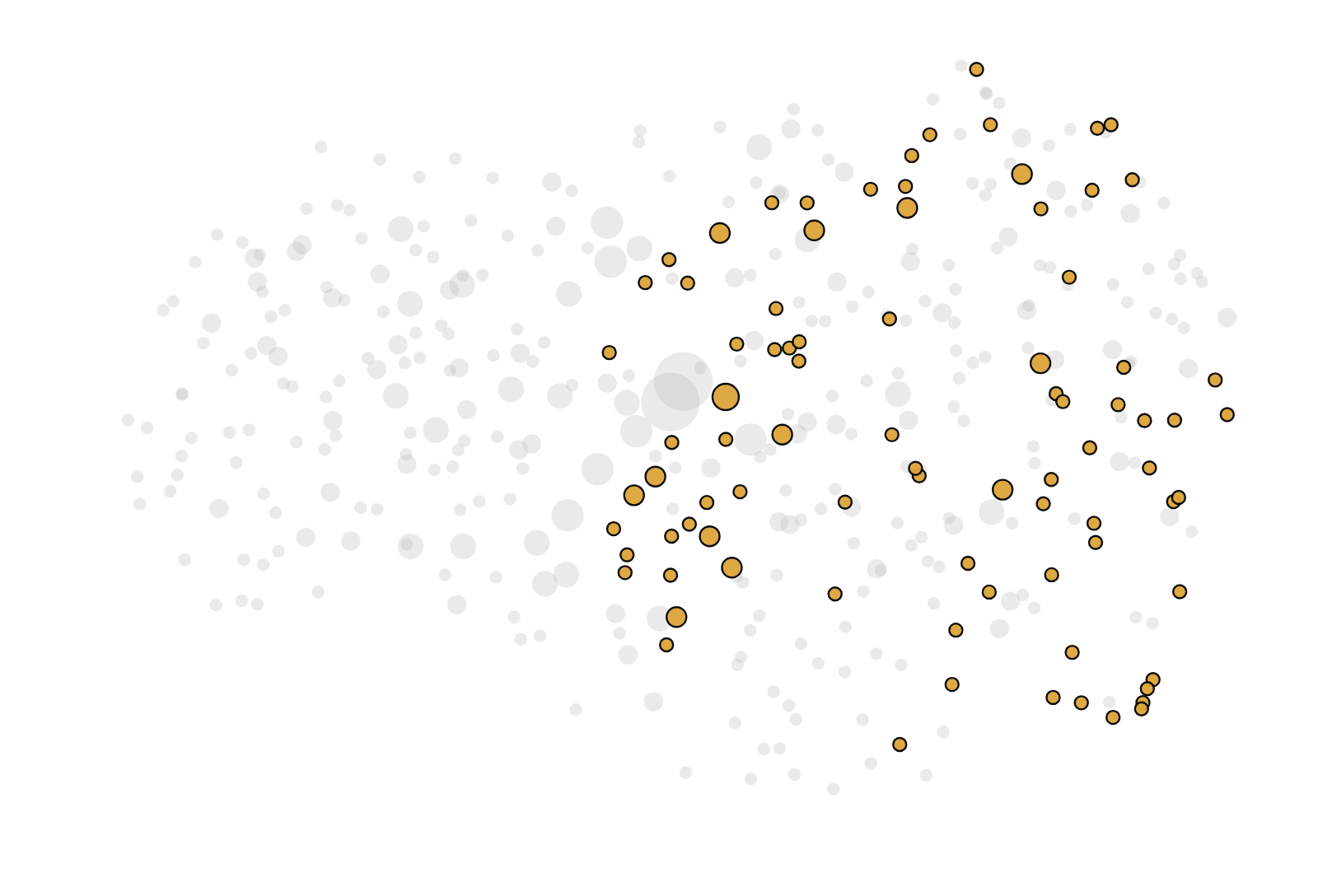}
            \caption{SDG 2: Zero Hunger\\(886, 1.8\%)}
        \end{subfigure}
                
        \begin{subfigure}{\textwidth}
        \includegraphics[width=\textwidth]{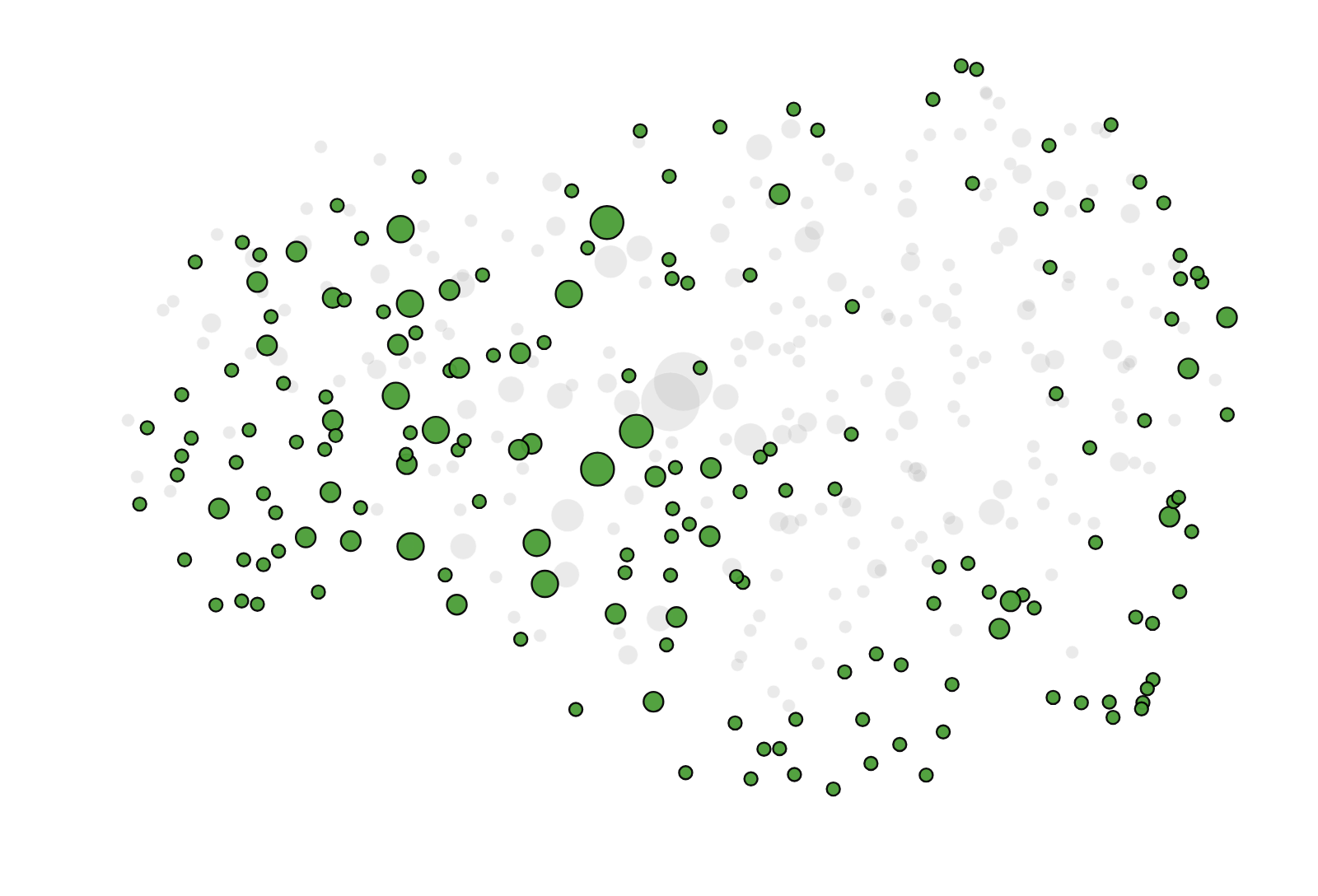}
        \caption{SDG 3: Good Health and Well-Being (9484, 19.1\%)}
        \end{subfigure}
    \end{minipage}

    
    \begin{subfigure}[b]{0.32\textwidth}
        \includegraphics[width=\textwidth]{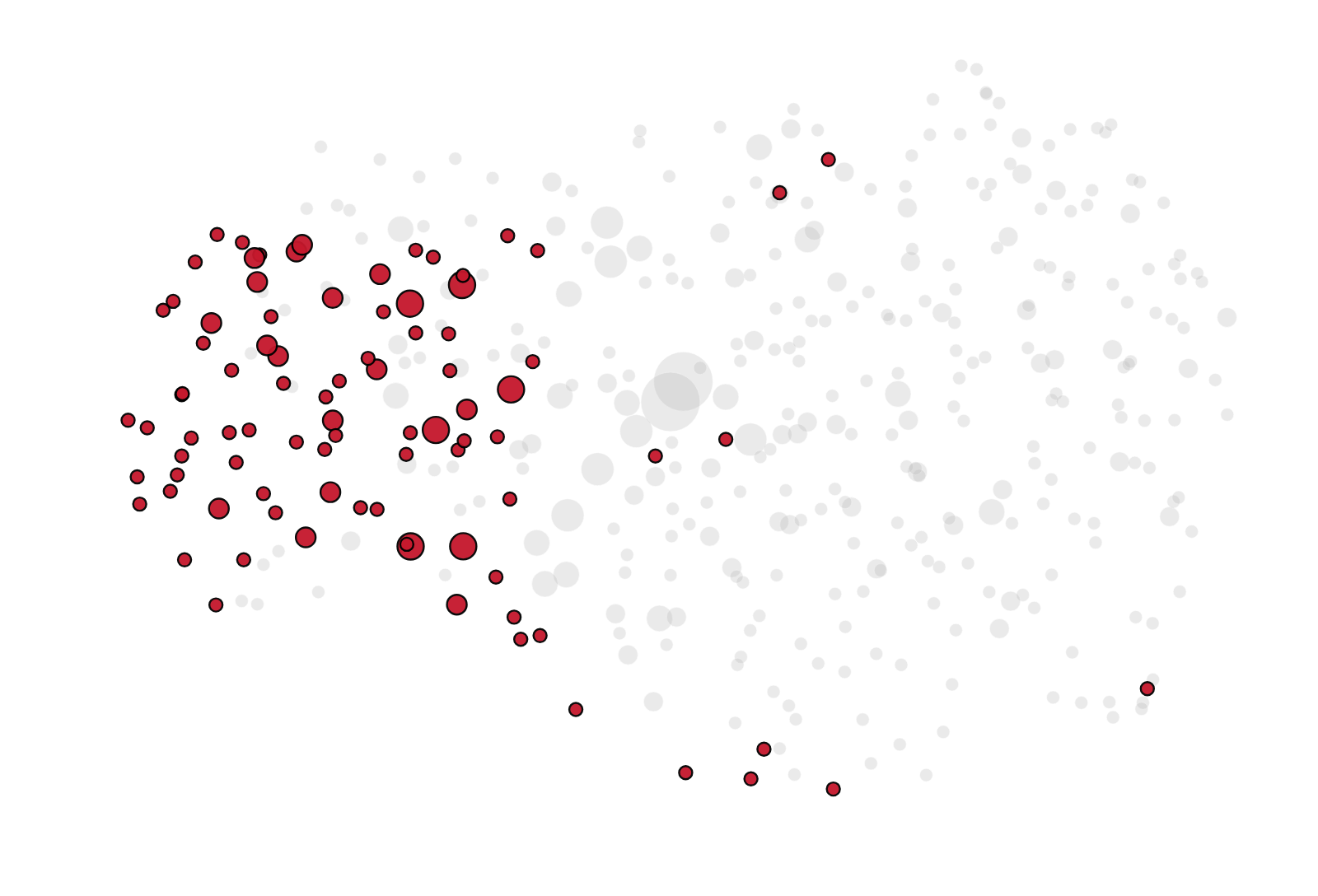}
        \caption{SDG 4: Quality Education\\(1951, 3.9\%)}
    \end{subfigure}
    \hfill
    \begin{subfigure}[b]{0.32\textwidth}
        \includegraphics[width=\textwidth]{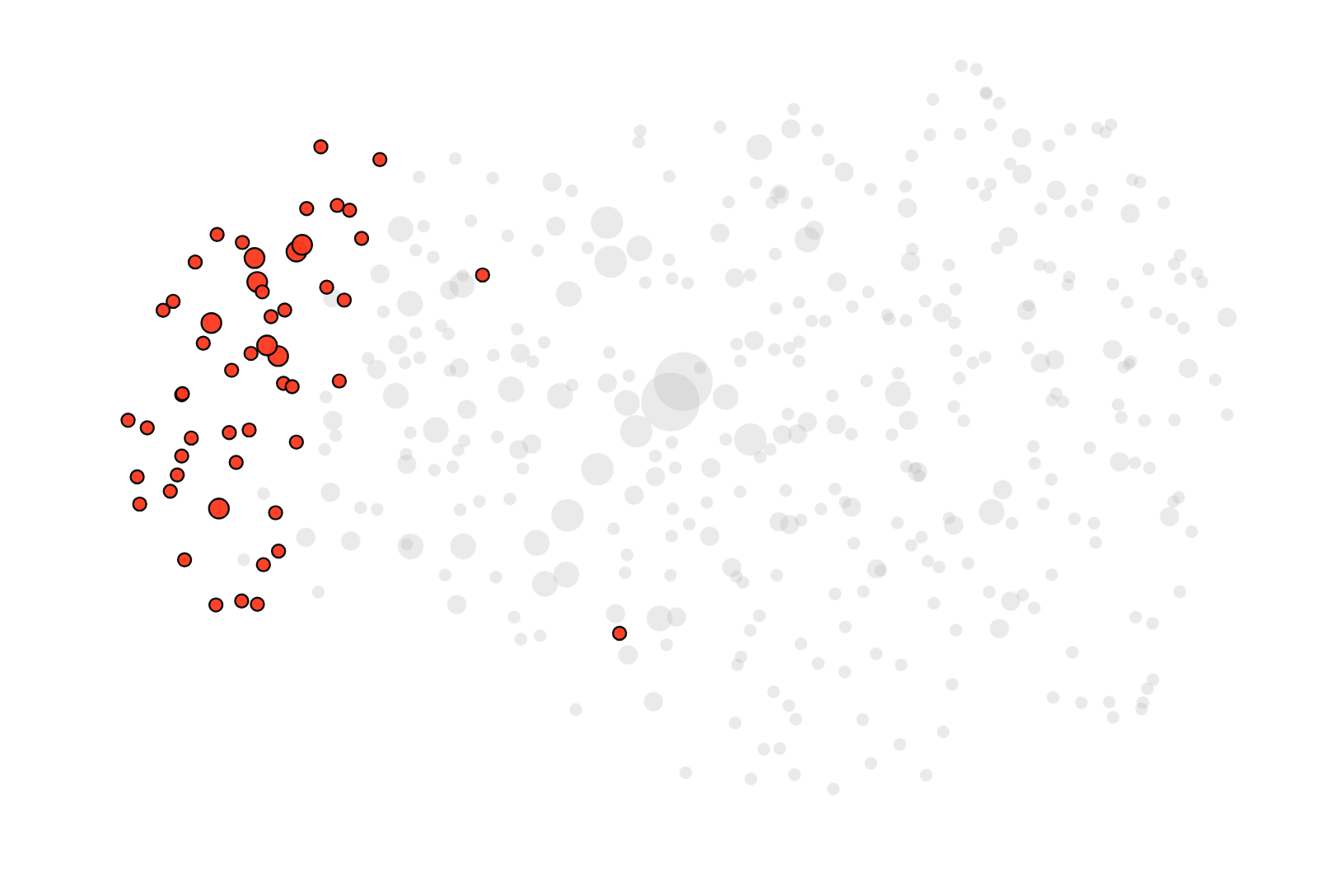}
        \caption{SDG 5: Gender Equality\\(290, 0.6\%)}
    \end{subfigure}
    \hfill
    \begin{subfigure}[b]{0.32\textwidth}
        \includegraphics[width=\textwidth]{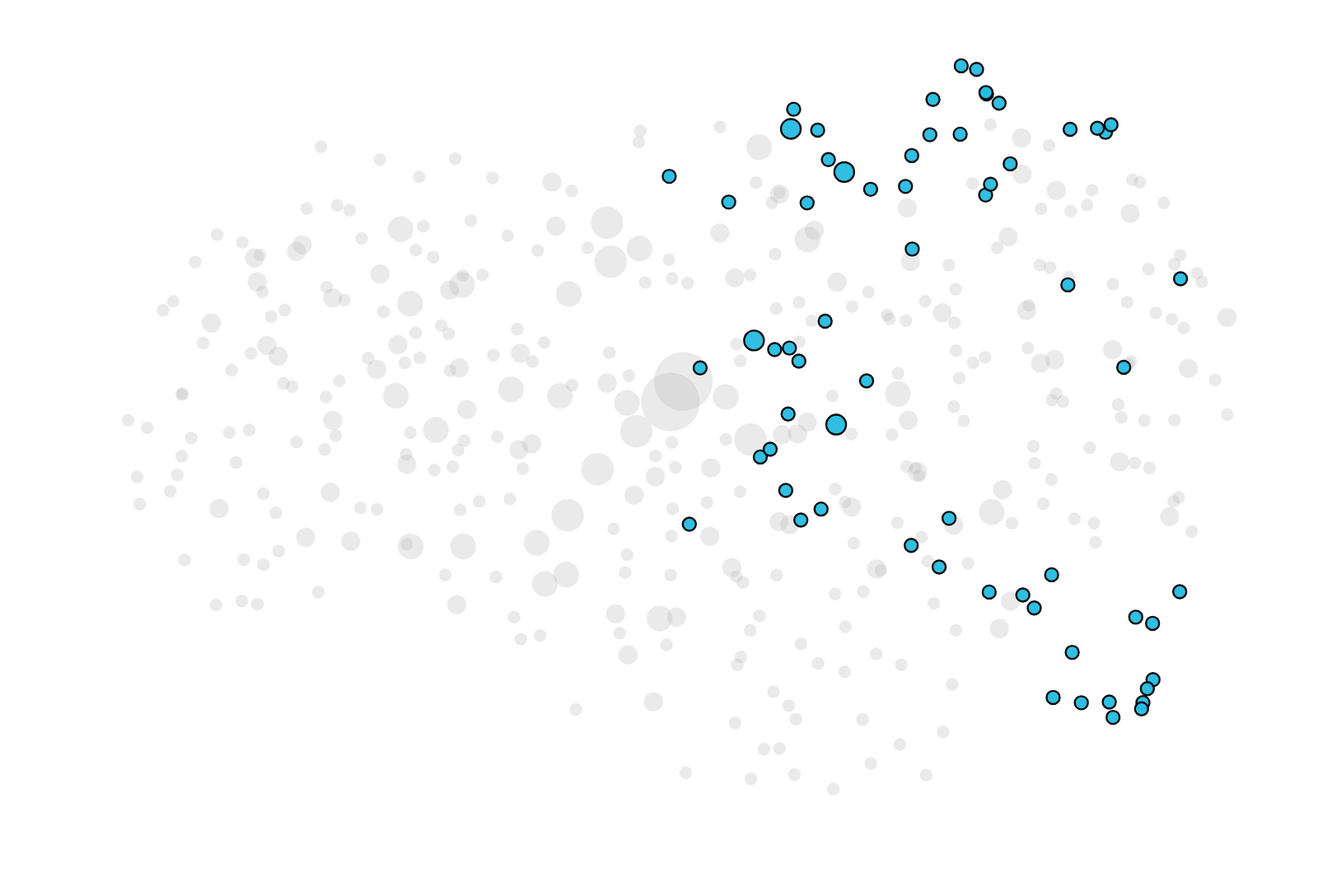}
        \caption{SDG 6: Clean Water \& Sanitation (165, 0.3\%)}
    \end{subfigure}

    \begin{subfigure}[b]{0.32\textwidth}
        \includegraphics[width=\textwidth]{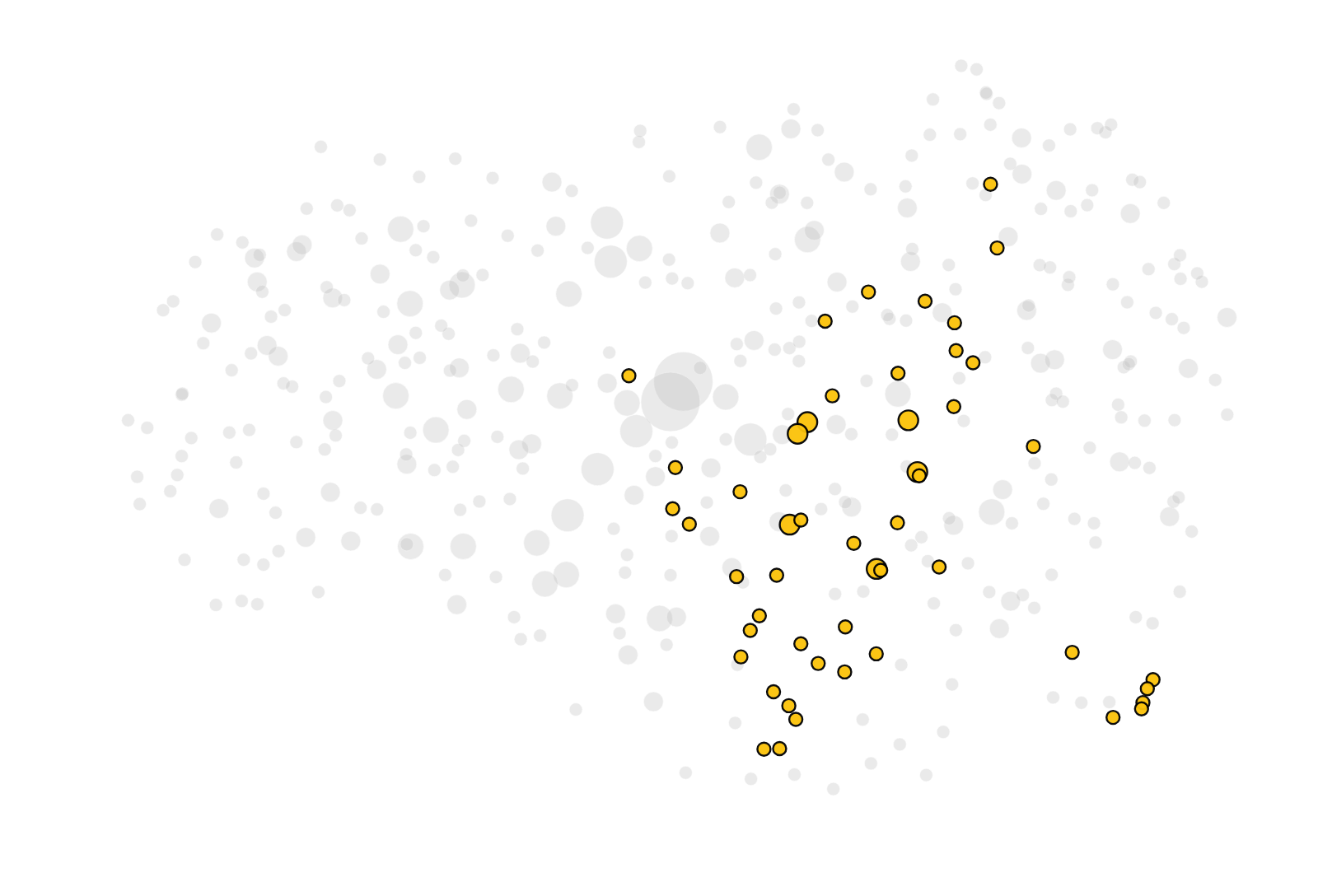}
        \caption{SDG 7: Affordable and Clean Energy (456, 0.9\%)}
    \end{subfigure}
    \hfill
    \begin{subfigure}[b]{0.32\textwidth}
        \includegraphics[width=\textwidth]{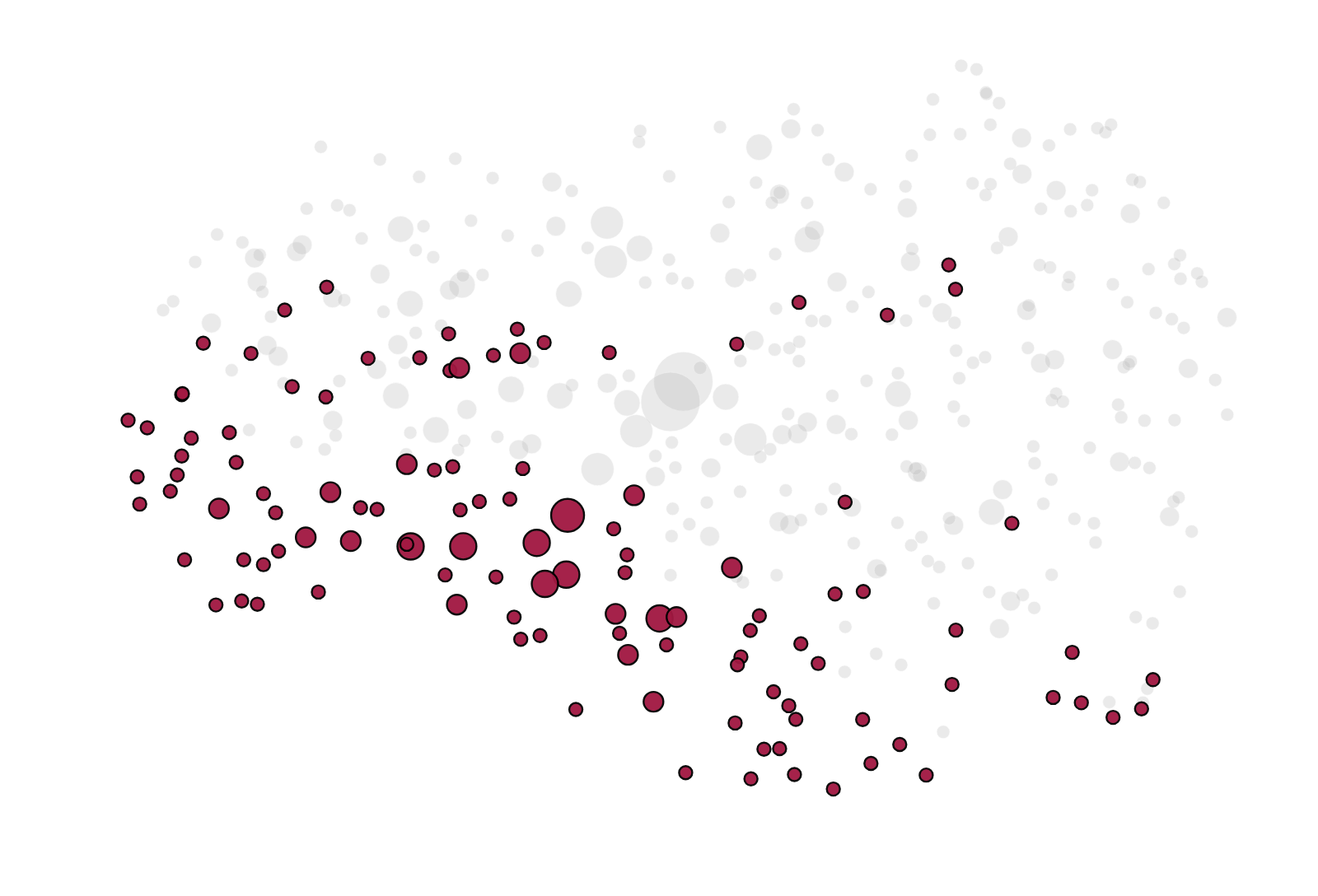}
        \caption{SDG 8: Decent Work\\(5286, 10.6\%)}
    \end{subfigure}
    \hfill
    \begin{subfigure}[b]{0.32\textwidth}
        \includegraphics[width=\textwidth]{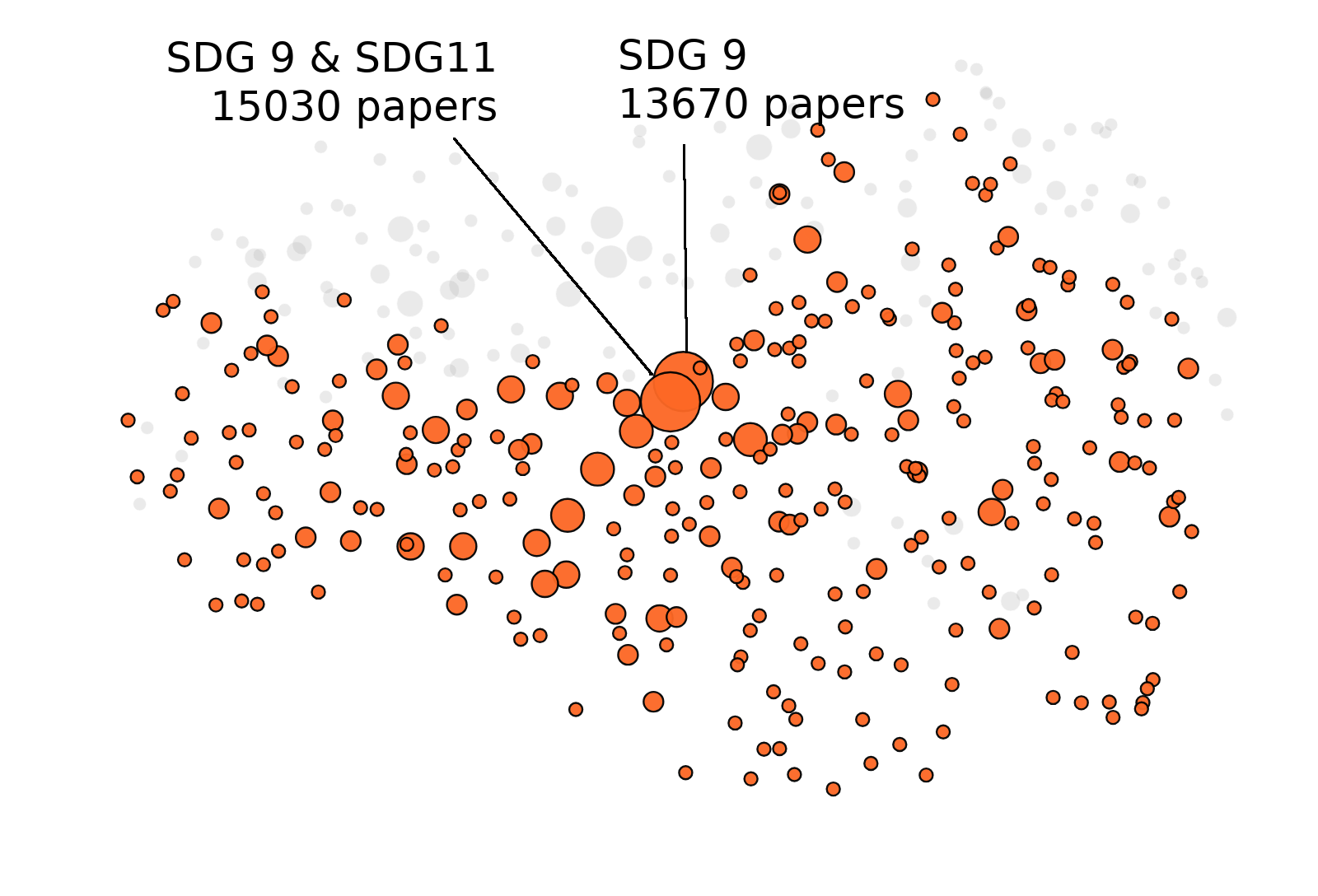}
        \caption{SDG 9: Industry\\(43521, 87.5\%)}
    \end{subfigure}

    \caption{The figures shows the 2D semantic projection of the 48,163 
    papers based on their SDG Alignment. Each node on the graph represents 
    a unique combination of SDGs, and its size corresponds to the number 
    of papers sharing that specific combination. Colored overlays indicate
    the distribution of papers aligned to each specific goal (SDGs 2 through 9), 
    alongside their absolute count and relative percentage. This visualization 
    demonstrates a massive centralization of research: the overwhelming majority 
    of papers are heavily concentrated in the center of the landscape,
    dominated by SDG 9 and SDG 11: almost 30,000 papers are
    concentrated in two nodes labeled as aligned with SDG 9 and SDG 9 \& SDG 11 combinations. In contrast, 
    research addressing other sustainability challenges is widely scattered across 
    the periphery in much lower-volume clusters. Note: SDG 1 is excluded 
    due to insufficient data.}
\label{fig:full_sdg_dashboard_portrait_p1}
\end{figure*}

\begin{figure*}[!htb] 
    \centering
    
    \begin{minipage}[c]{0.64\textwidth}
        \begin{subfigure}{\textwidth}
            \includegraphics[width=\textwidth]{figures/banner/banner.png}
            \caption{\textbf{All Papers (49,751, 100\%)}}
        \end{subfigure}
    \end{minipage}
    \hfill
    \begin{minipage}[c]{0.32\textwidth}
        \begin{subfigure}{\textwidth}
        \includegraphics[width=\textwidth]{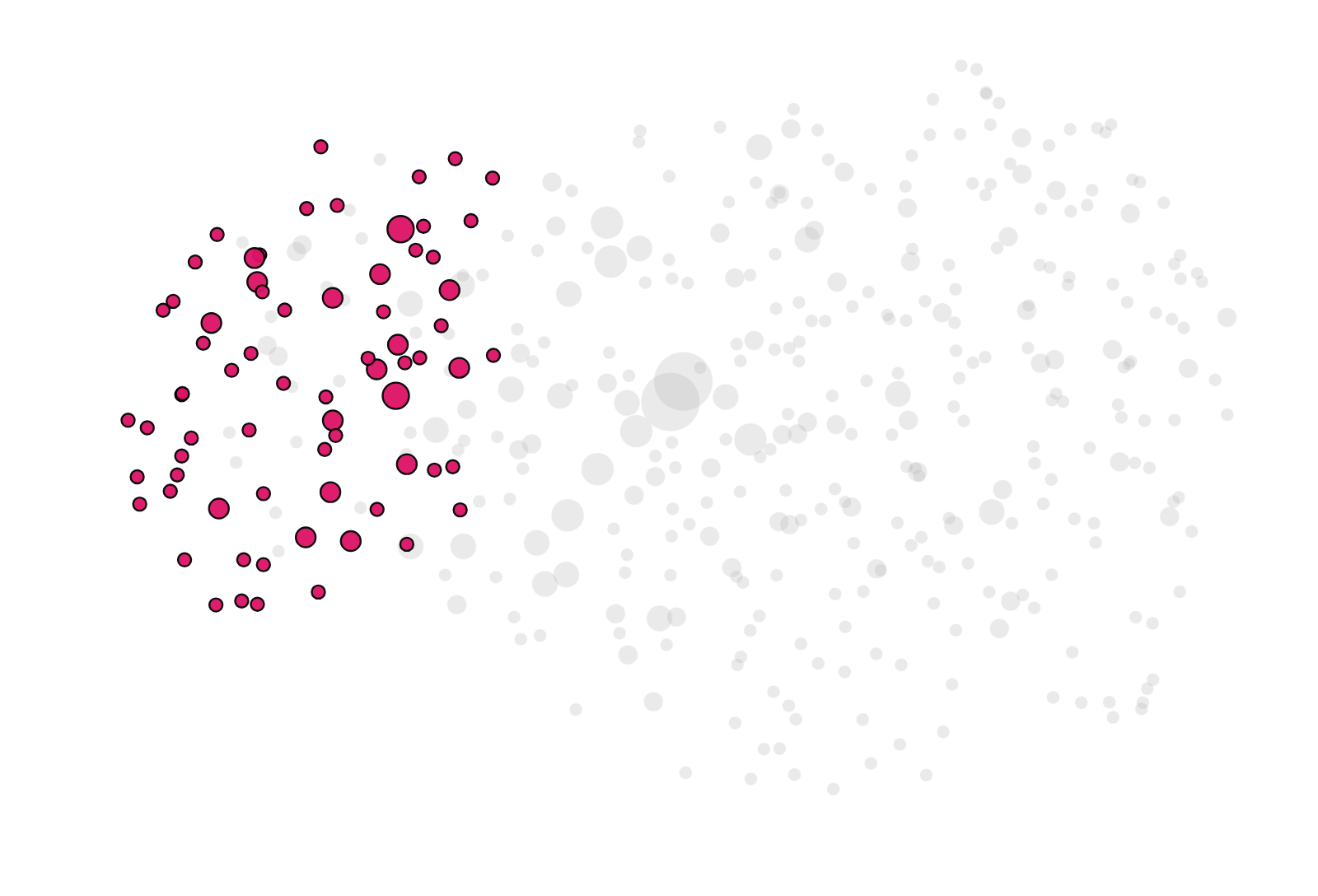}
        \caption{SDG 10: Reduced Inequalities\\(988, 2.0\%)}
        \end{subfigure}
        
        \vspace{0.4cm}
        
        \begin{subfigure}{\textwidth}
        \includegraphics[width=\textwidth]{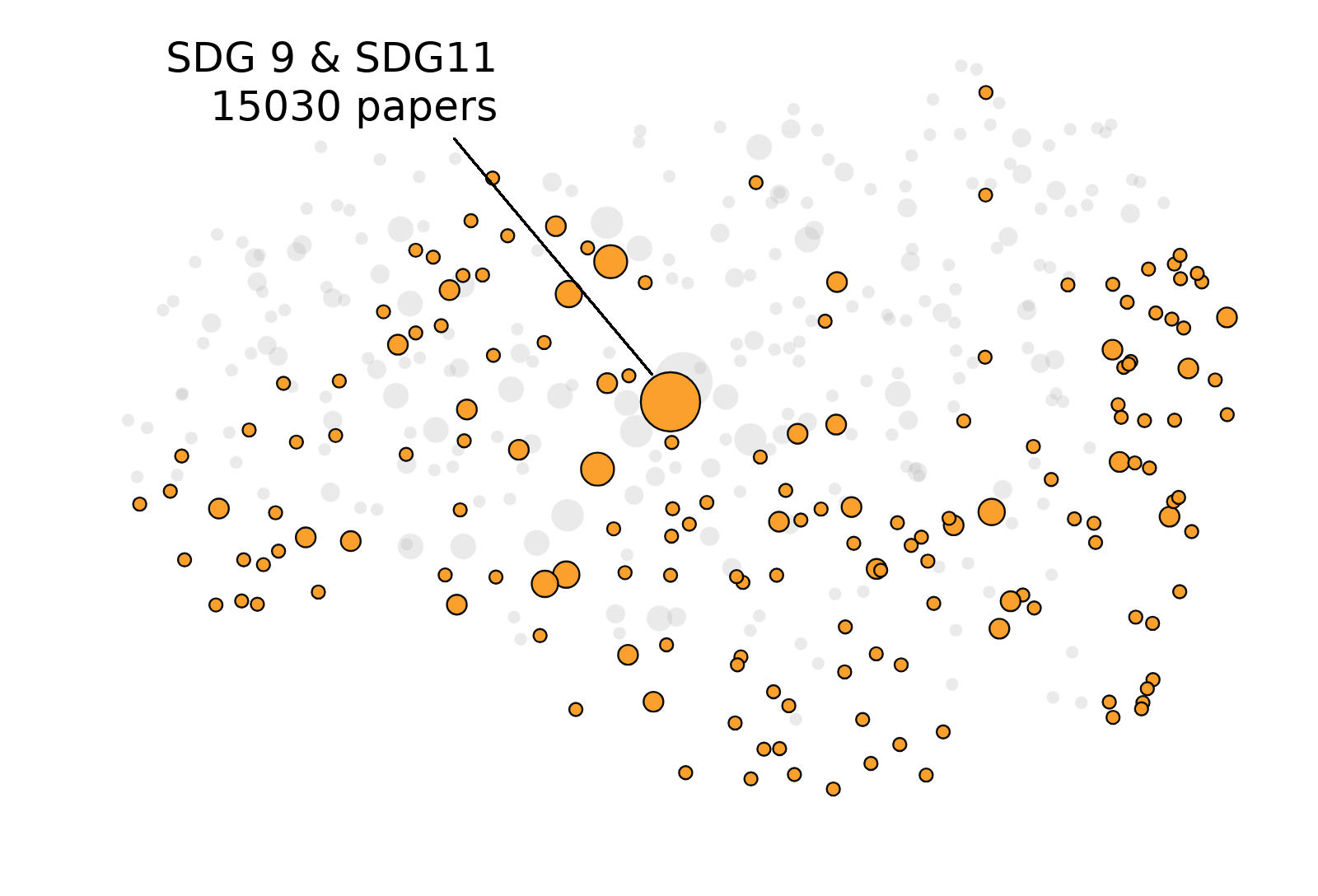}
        \caption{SDG 11: Sustainable Cities\\(21642, 43.5\%)}
        \end{subfigure}
    \end{minipage}
    
    \begin{subfigure}[b]{0.32\textwidth}
        \includegraphics[width=\textwidth]{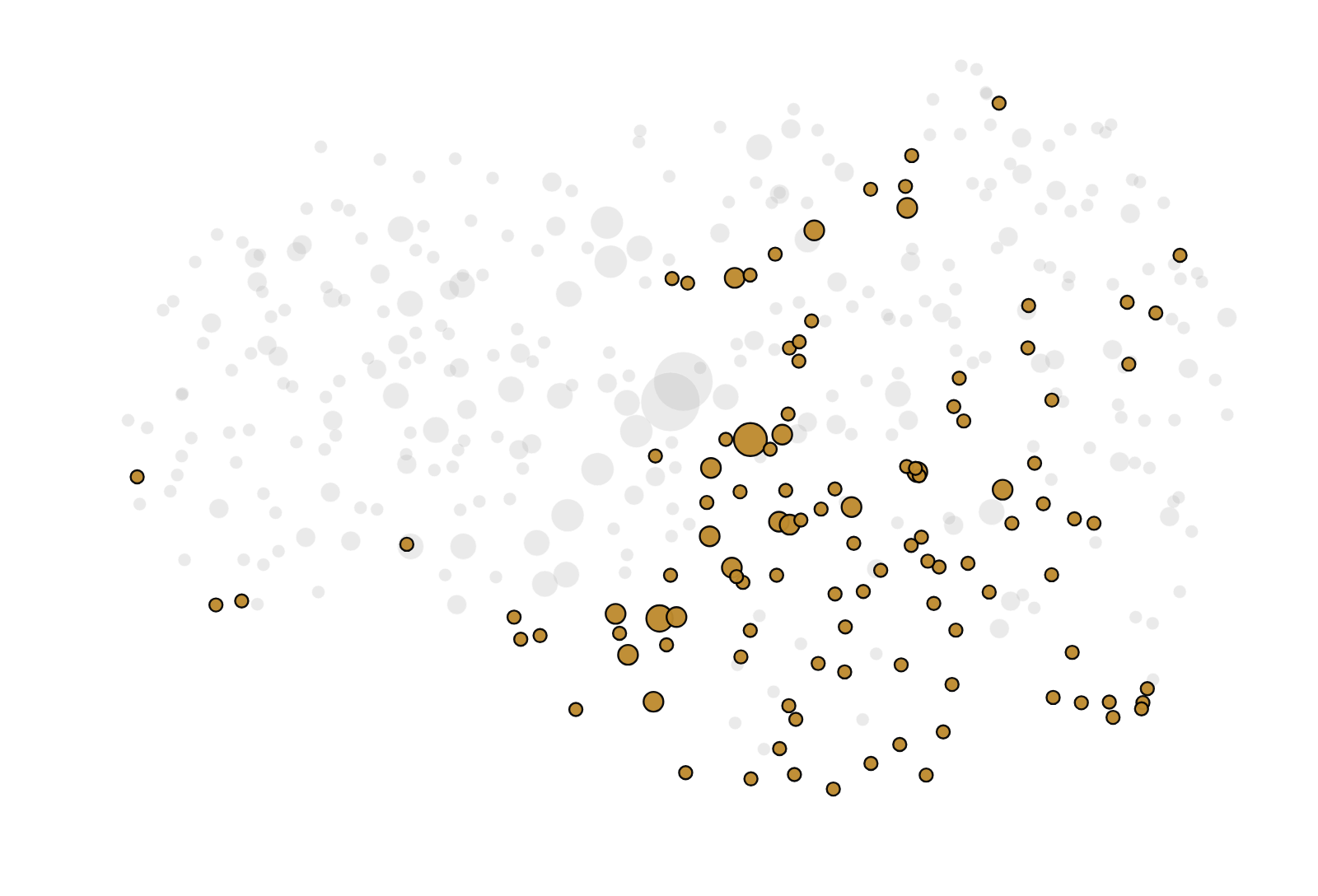}
        \caption{SDG 12: Responsible Production (2410, 4.8\%)}
    \end{subfigure}
    \hfill
    \begin{subfigure}[b]{0.32\textwidth}
        \includegraphics[width=\textwidth]{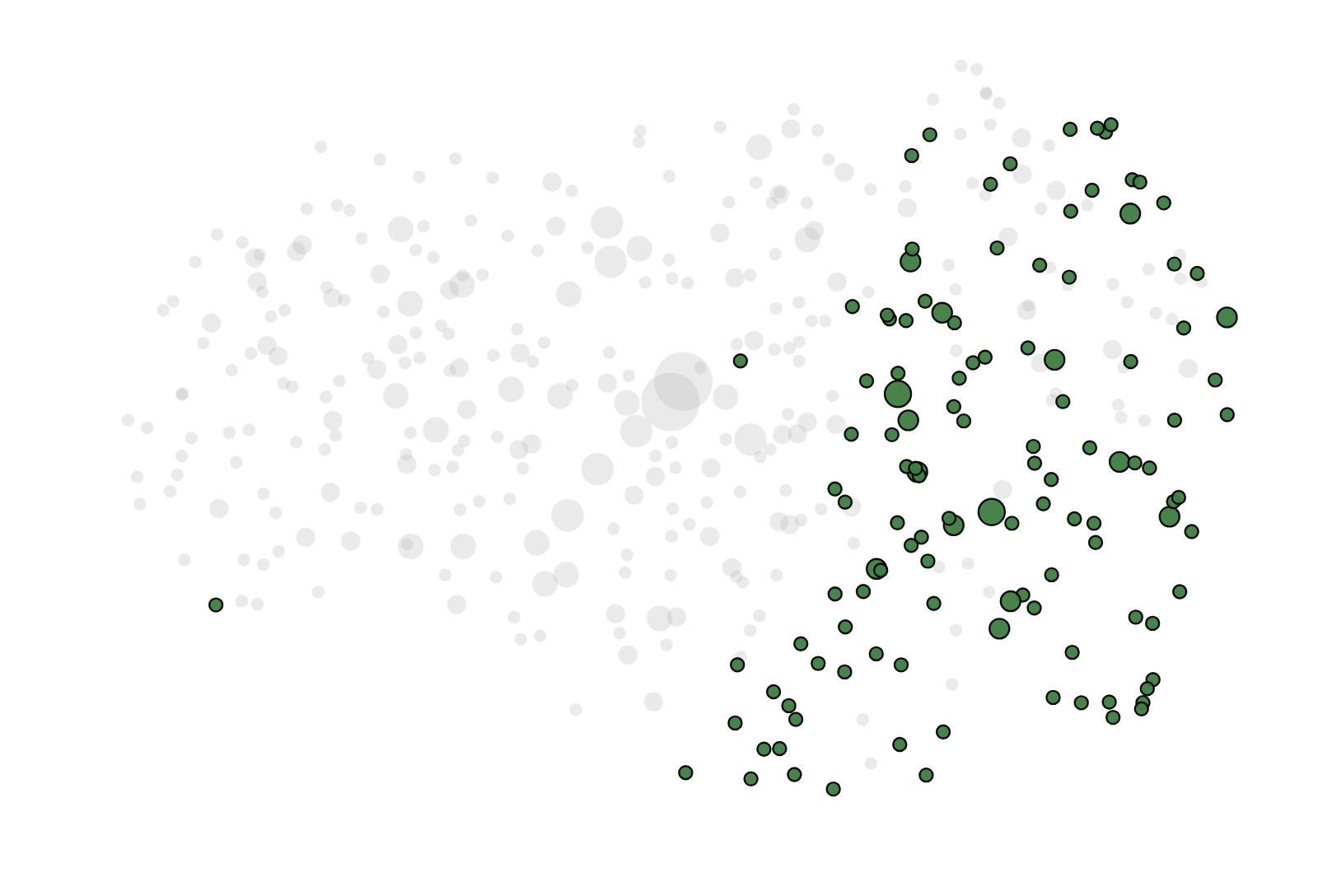}
        \caption{SDG 13: Climate Action\\(1199, 2.4\%)}
    \end{subfigure}
    \hfill
    \begin{subfigure}[b]{0.32\textwidth}
        \includegraphics[width=\textwidth]{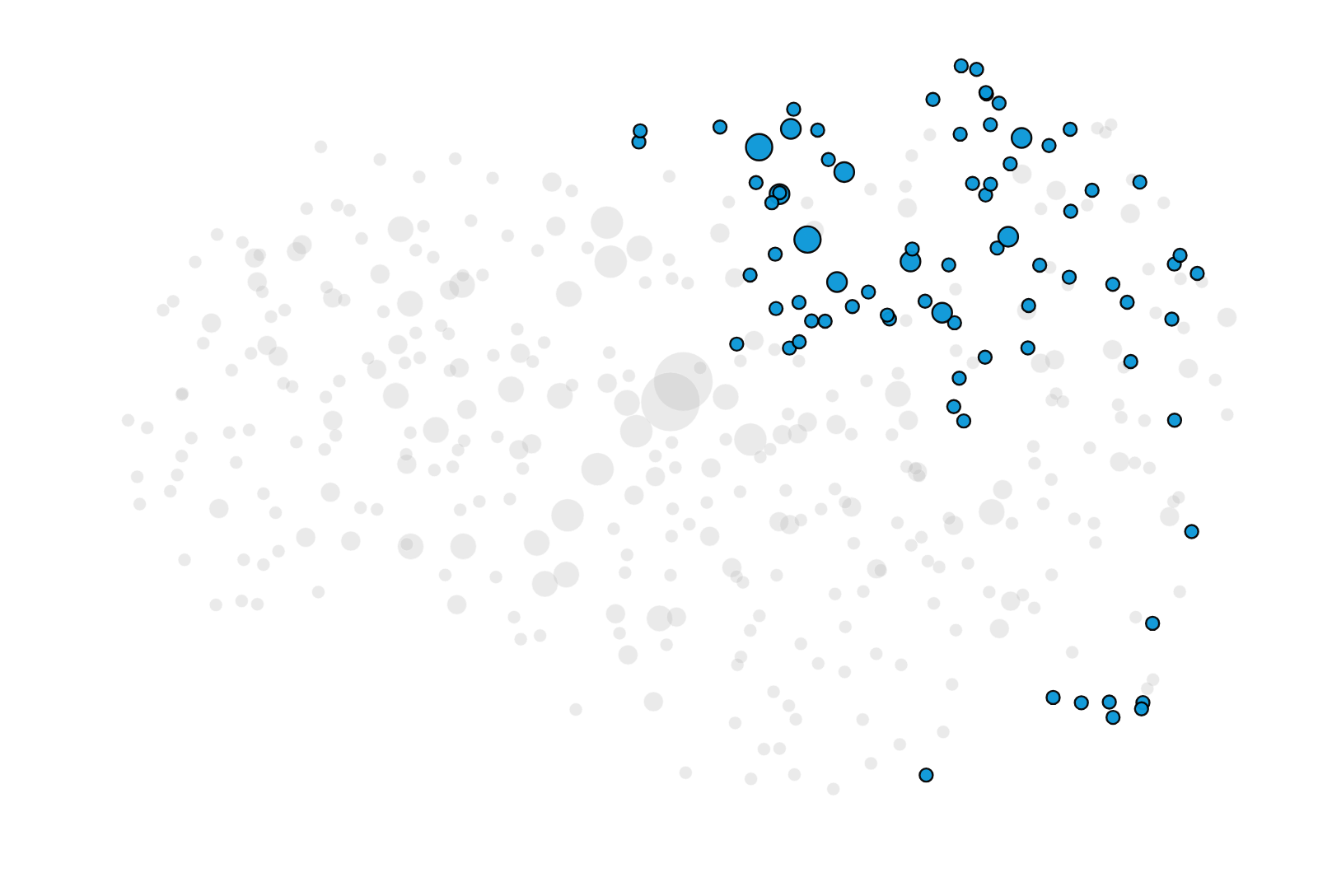}
        \caption{SDG 14: Life Below Water \\(1278, 2.6\%)}
    \end{subfigure}

    \vspace{0.4cm}

    \begin{subfigure}[b]{0.32\textwidth}
        \includegraphics[width=\textwidth]{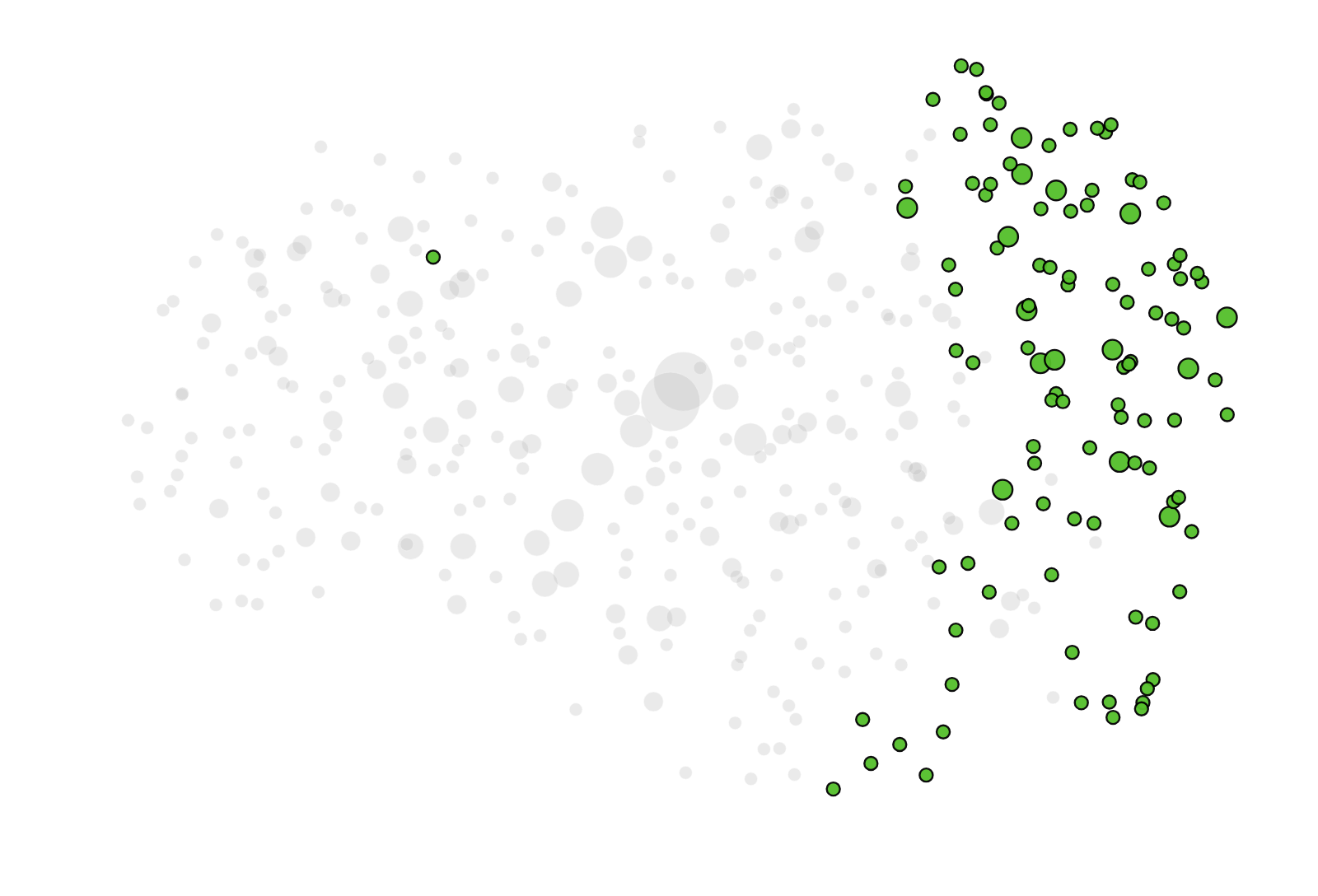}
        \caption{SDG 15: Life On Land\\(777, 1.6\%)}
    \end{subfigure}
    \hfill
    \begin{subfigure}[b]{0.32\textwidth}
        \includegraphics[width=\textwidth]{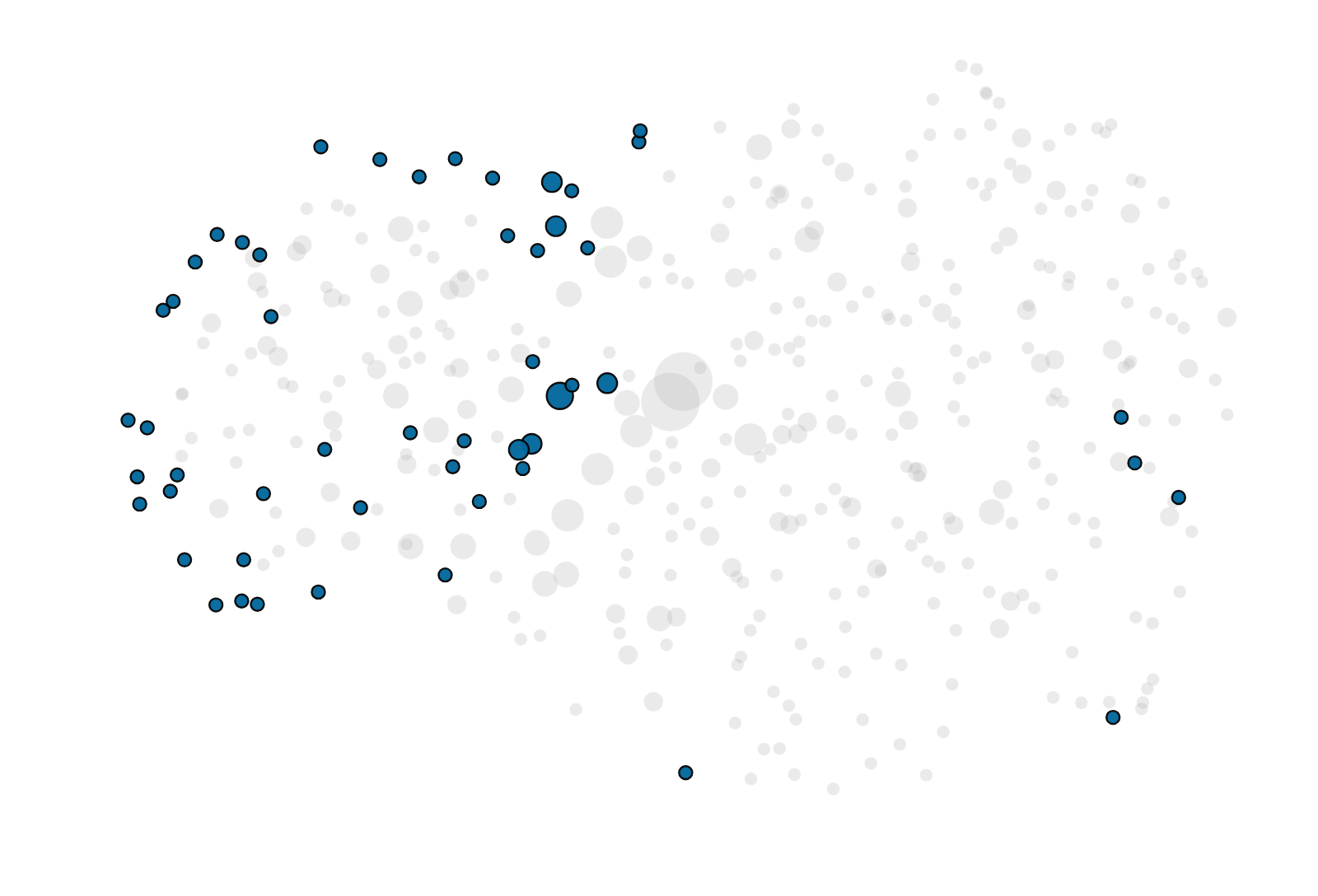}
        \caption{SDG 16: Peace, Justice \& Strong Institutions (355, 0.7\%)}
    \end{subfigure}
    \hfill
    \begin{subfigure}[b]{0.32\textwidth}
        \includegraphics[width=\textwidth]{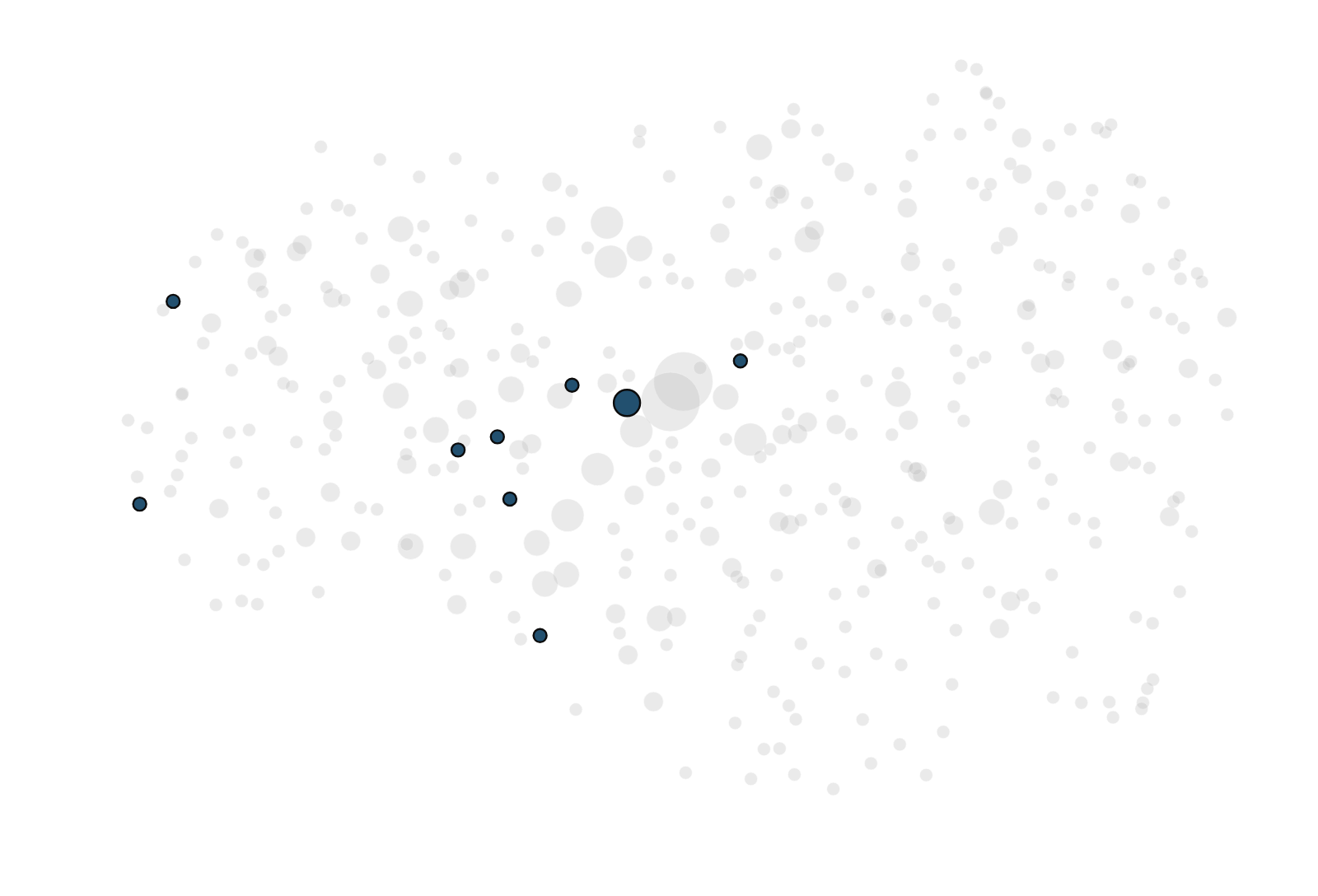}
        \caption{SDG 17: Partnerships for the Goals (159, 0.3\%)}
    \end{subfigure}

    \vspace{0.5cm}
    \caption{Semantic landscape of robotics research across UN Sustainable Development 
    Goals (Part 2 of 2). Continuing the global corpus projection from Part 1, colored 
    overlays indicate the distribution of papers aligned to goals 10 through 17. 
    As established, node size corresponds to the volume of papers sharing a 
    specific SDG combination. This mapping visually reinforces the centralised 
    structure of the field: the massive central hub is dominated by 
    infrastructural goals, most notably SDG 11 (Sustainable Cities). 
    The ecological and climate-focused research, 
    with Climate Action (SDG 13), Life Below Water (SDG 14), and 
    Life on Land (SDG 15) is divided into more isolated, and low-volume 
    clusters scattered across around the center.}
    \label{fig:full_sdg_dashboard_portrait_p2}
\end{figure*}

\clearpage
\section{Appendix: Cross-Model Classification Benchmarking}
\label{sec:model_comp}
To validate the robustness of the zero-shot classification pipeline and mitigate the risk of 
single-model bias, we conducted a comparative ensemble analysis. A subset of the first 500 papers 
published in 2026 was processed using identical system and paper-level prompts across
four different Large Language Models:  \texttt{DeepSeek-V3} (used in the main analysis), \texttt{Qwen3-32B},
\texttt{Llama-3.1-70B-Instruct}, and \texttt{gpt-oss-20b}. 

\texttt{DeepSeek-V3} was inferred using the DeepSeek API (as in the main analysis), 
while the other three models were inferred through the HuggingFace Inference API, ensuring consistent input formatting
and response parsing and all the models are publicly available on HuggingFace Model Hub.

\begin{figure}[!htb]
    \centering
    \includegraphics[width=0.9\textwidth]{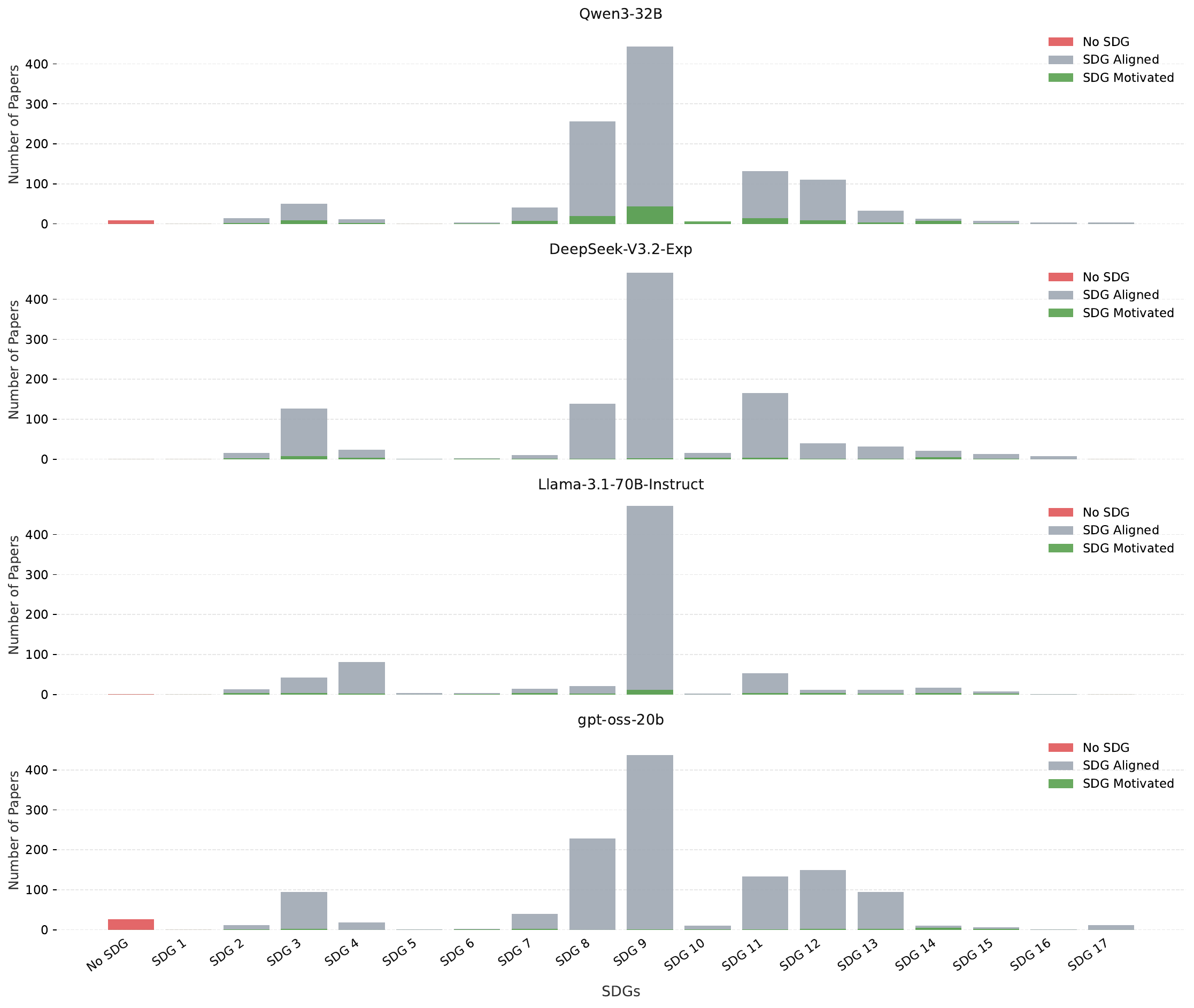}
    \caption{Small multiples bar chart comparing the distribution of papers across all 17 SDGs by model. 
    Each chart displays the absolute number of papers classified as having No SDG relevance, being SDG Aligned
    or being explicitly SDG Motivated. The shared axes highlight the consistent distribution shape 
    across all four models.}
    \label{fig:anx_sdg_papers}
\end{figure}

As demonstrated in \Cref{fig:anx_sdg_papers}, while the variations in SDG Aligned and SDG Motivated 
classification exist due to different underlying architectures, the macro-level trends remain 
relatively consistent. SDG 9 (Industry, Innovation, and Infrastructure) dominates the SDG Aligned category across 
all models, followed by SDG 11 (Sustainable Cities and Communities) and SDG 8 (Decent Work and Economic Growth) 
and SDG 3 (Good Health and Well-being). 
The distribution of other SDGs is more variable, but overall very low across all models, 
with SDGs 1, 2, 5, 6, 10, 14, 15 and 16 consistently appearing in the lowest volume clusters.
Interestingly, all the models show the similar optimistic bias towards classifying papers as SDG Aligned papers,
with almost all papers being classified as aligned to at least one SDG. 

The amount of papers classified as explicitly motivated by sustainability (SDG Motivated) 
is consistently low across all models. However, it is noticeable that \texttt{Qwen3-32B} is
the most optimistic model, classifying much more papers as SDG Motivated, while \texttt{DeepSeek-V3} seems to be 
the most conservative one.

\begin{figure}[!htb]
    \centering
    
    \centering
    \includegraphics[width=0.7\textwidth]{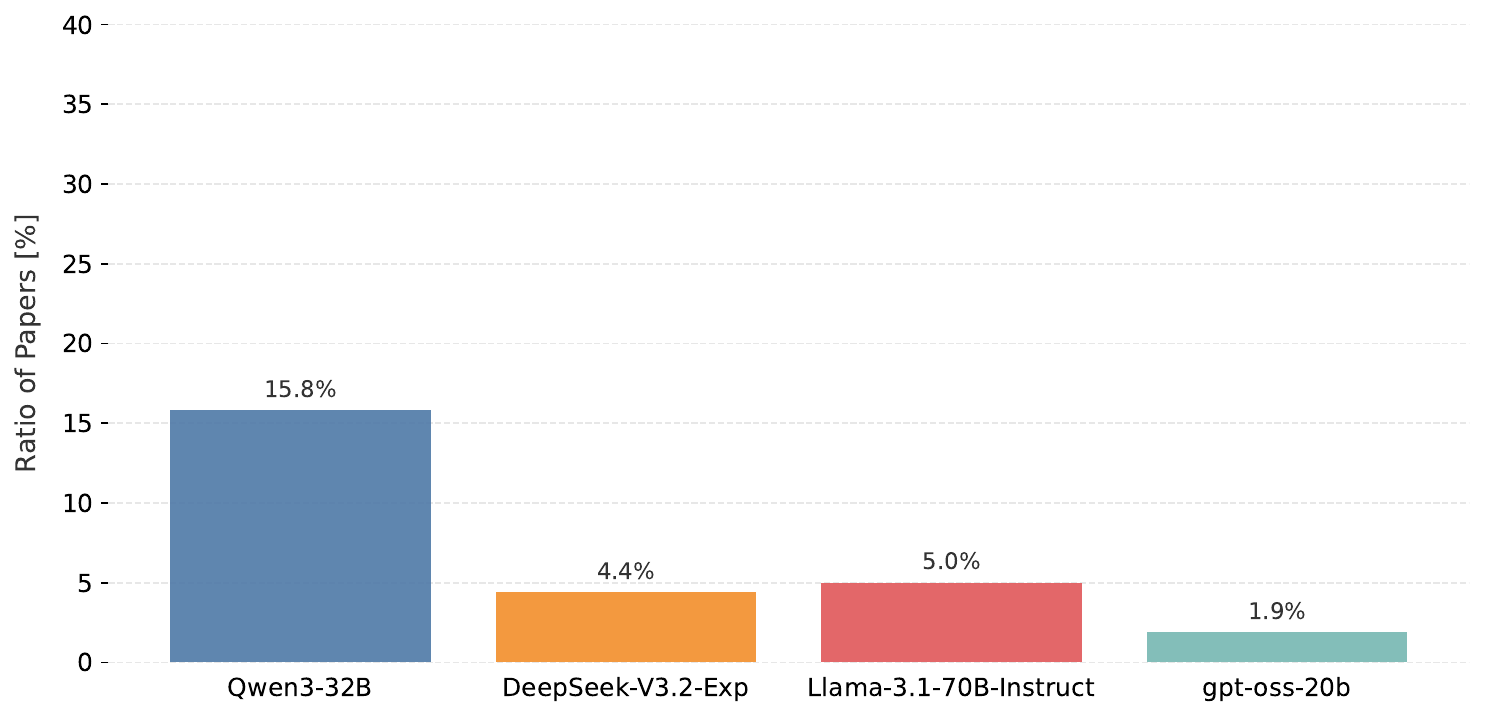}
    \caption{Percentage of papers classified as explicitly motivated by sustainability (SDG Motivated) across all SDGs, for 
    4 tested models.}
    \label{fig:anx_sustainability_rate}
\end{figure}

\Cref{fig:anx_sustainability_rate} shows the average percentage of papers identified as explicitly motivated by 
sustainability-related topics per model. These results confirm the previous finding of a low explicit 
sustainability motivation in the field, with all models reporting low rate of SDG Motivated papers. 
Interestingly, \texttt{Qwen3-32B} is the most optimistic model, classifying 15\% papers as SDG Motivated, 
while \texttt{gpt-oss-20b} is the most pessimistic one with only 2\% of papers classified as SDG Motivated.

\begin{figure}[!htb]
    \centering
    \includegraphics[width=0.9\textwidth]{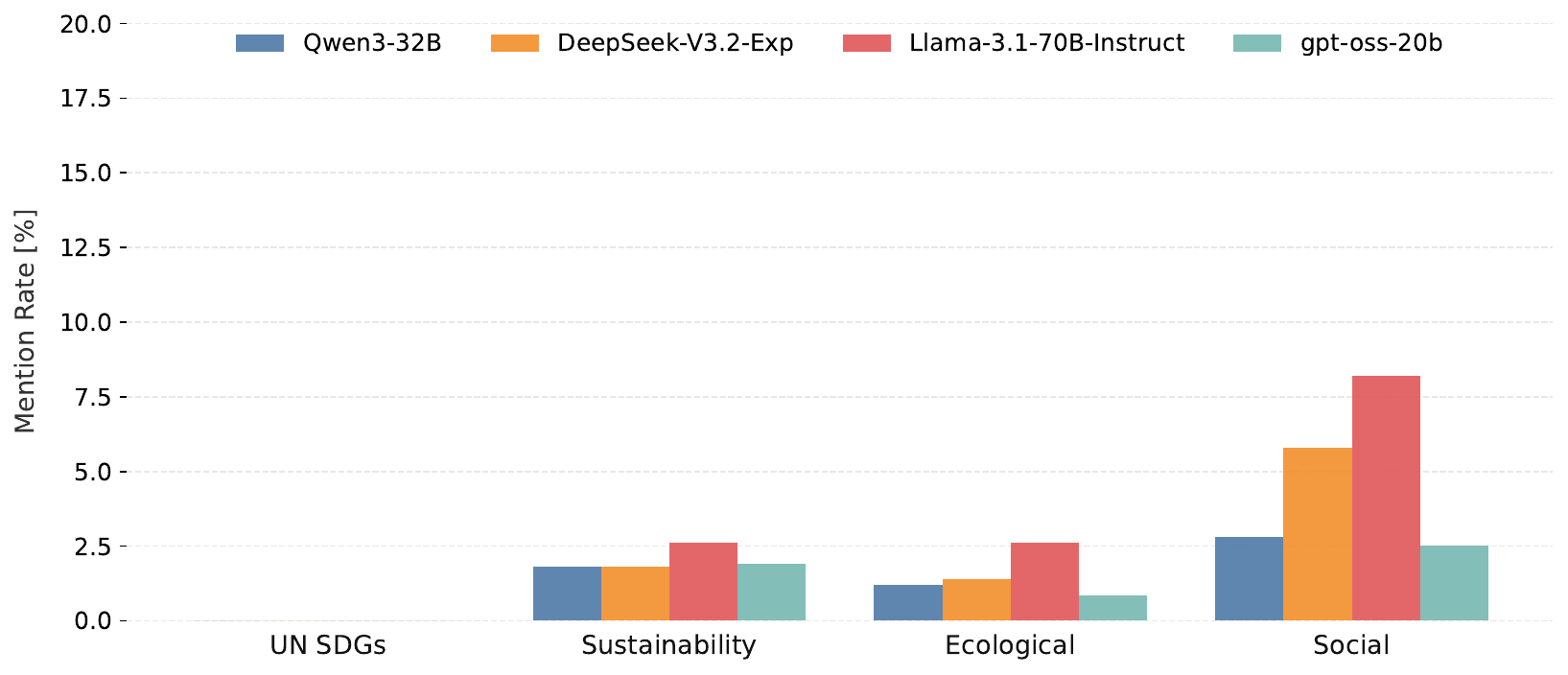}
    \caption{The chart 
    shows the percentage of papers identified as explicitly mentioning the UN SDGs, 
    Sustainability, Ecological impact, and Social impact for each LLM. While minor variations in 
    sensitivity exist (e.g., \texttt{Llama-3.1-70B-Instruct} and \texttt{DeepSeek-V3} extracting slightly more Social impact mentions), 
    the relative hierarchy and the near-total absence of UN SDG mentions remain 
    consistent.}
    \label{fig:anx_impact_mentions}
\end{figure}

When comparing the models' ability to extract mentions of key sustainability-related terms, 
we see that all the models show a similar pattern, with the highest mention rate for "Social impact", 
followed by "Ecological impact" and "Sustainability". The mention rate for UN SDGs is zero 
across all models, confirming the previous finding of a very low explicit sustainability motivation in the field.
In this case the most optimistic model is \texttt{Llama-3.1-70B-Instruct}, followed by \texttt{DeepSeek-V3} 
which extracts more "Social impact" mentions than the other models, but the results are within 
the same order of magnitude as the other models. 

\subsection*{Analysis Footprint}

The ensemble analysis was conducted on a subset of 500 papers (+ 100 for testing), which were processed sequentially across all four models.
The total number of API calls made was 2400 (600 papers x 4 models), with the average prompt size of around 25,000 
tokens per paper (including the system prompt and the paper-level prompt). 

The analysis from~\citep{jegham2025hungry} suggests the values reported in \Cref{tab:energy_footprint}:

\begin{table}[!htb]
    \centering
    \small
    \caption{Carbon Intensity of Providers (CIP) and energy consumption per prompt for models with available data~\citep{jegham2025hungry}.}
    \label{tab:energy_footprint}
    \begin{tabular}{lcc}
        \toprule
        \textbf{Model} & \textbf{CIP} & \textbf{Energy per prompt (Wh)} \\
        \midrule
        \texttt{DeepSeek-V3}       & 0.600 & 13.162 \\
        \texttt{Llama-3.1-70B}     & 0.287 & 19.183 \\
        \bottomrule
    \end{tabular}
\end{table}
The other two models' energy consumption is not publicly available. Both of these models are 
most likely more energy efficient than both \texttt{Llama-3.1-70B} and \texttt{DeepSeek-V3},
as they are smaller in size. 

However, if we consider the worst case scenario, taking the highest energy consumption per prompt 19.183Wh 
for both \texttt{Qwen3-32B} and \texttt{gpt-oss-20b}, the total energy consumption of the analysis could
be estimated as 42.3kWh.
\begin{equation}
E = \underbrace{600 \times 13.162Wh}_{\texttt{DeepSeek-V3}} + 1800 \times 19.183Wh = 42.3kWh
\end{equation}

The same worst case scenario can be estimated by taking the CIP of 0.6 (corresponding to \texttt{DeepSeek-V3})
for both \texttt{Qwen3-32B} and \texttt{gpt-oss-20b}. The carbon footprint of this analysis would be around 21.9kgCO$_2$eq.
\begin{equation}
C = \underbrace{600 \times 0.013kWh \times 0.6}_{\texttt{DeepSeek-V3}} + 
\underbrace{1200 \times 0.019kWh \times 0.6}_{\texttt{Qwen3-32B}\text{ \& }\texttt{gpt-oss-20b}} +  
\underbrace{600 \times 0.019kWh \times 0.287}_{\texttt{Llama-3.1-70B}} 
= 21.853kgCO_2eq
\end{equation}

Finally, the overall price of the analysis was around 35\$ (for 2,600 papers).

\section{Appendix: Consistency of DeepSeek-V3 classifications}
\label{sec:consistency}
To verify the robustness and repeatability of our classification pipeline, we conducted a consistency 
analysis by running DeepSeek-V3 model on two different papers, one representing a 
purely technical paper with no explicit sustainability motivation (Example A: \href{http://arxiv.org/abs/1501.06862}{arXiv:1501.06862})
and another representing a sustainability-driven paper (Example B: \href{https://arxiv.org/abs/2411.15159v1}{arXiv:2411.15159} ). 
The same system prompt and paper-level prompt were used for each run, 
and the model's classifications were recorded across the 200 runs for each paper.

At each run, we recorded the model's classification for the paper type, 
the SDGs and targets identified as explicitly motivated by sustainability, 
the SDGs and targets identified as relevant to the technologies or methods 
developed in the paper, the mentions of key sustainability-related terms, 
and the reasoning provided by the model.

\subsection{Consistency of impact mentions}

\begin{figure}[!htb]
    \centering
    \includegraphics[width=0.8\textwidth]{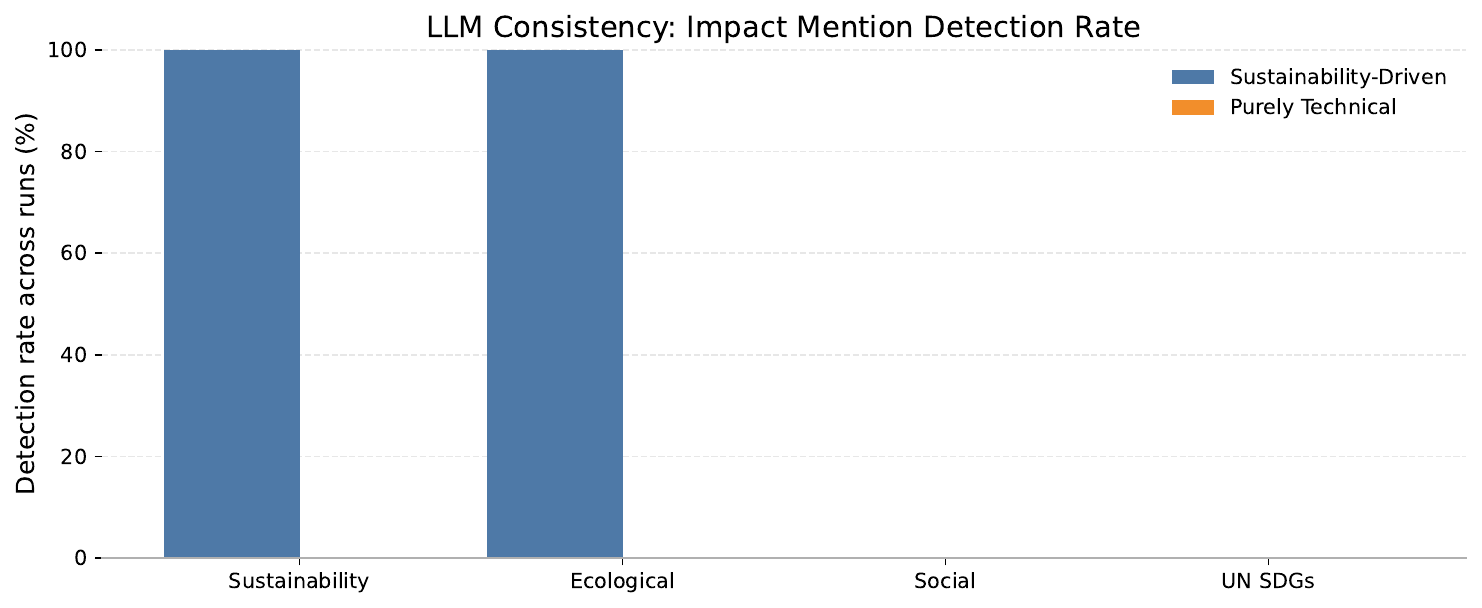}
    \caption{Ratio of classified impact mentions across 200 runs for both 
    Example A(Purely technical) and Example B (Sustainability-Driven).}
    \label{fig:consistency_impact_rates}
\end{figure}

The results of the impact mentions consistency analysis are shown in \Cref{fig:consistency_impact_rates}.
The sustainability-driver paper (Example B) is consistently classified as mentioning "Sustainability" 
and "Ecological impact", while the purely technical paper (Example A) is consistently 
classified as not mentioning any of the key sustainability-related terms.

\subsection{Consistency of SDG Motivation and Alignment}

\begin{figure}[!htb]
    \centering
    \includegraphics[width=\textwidth]{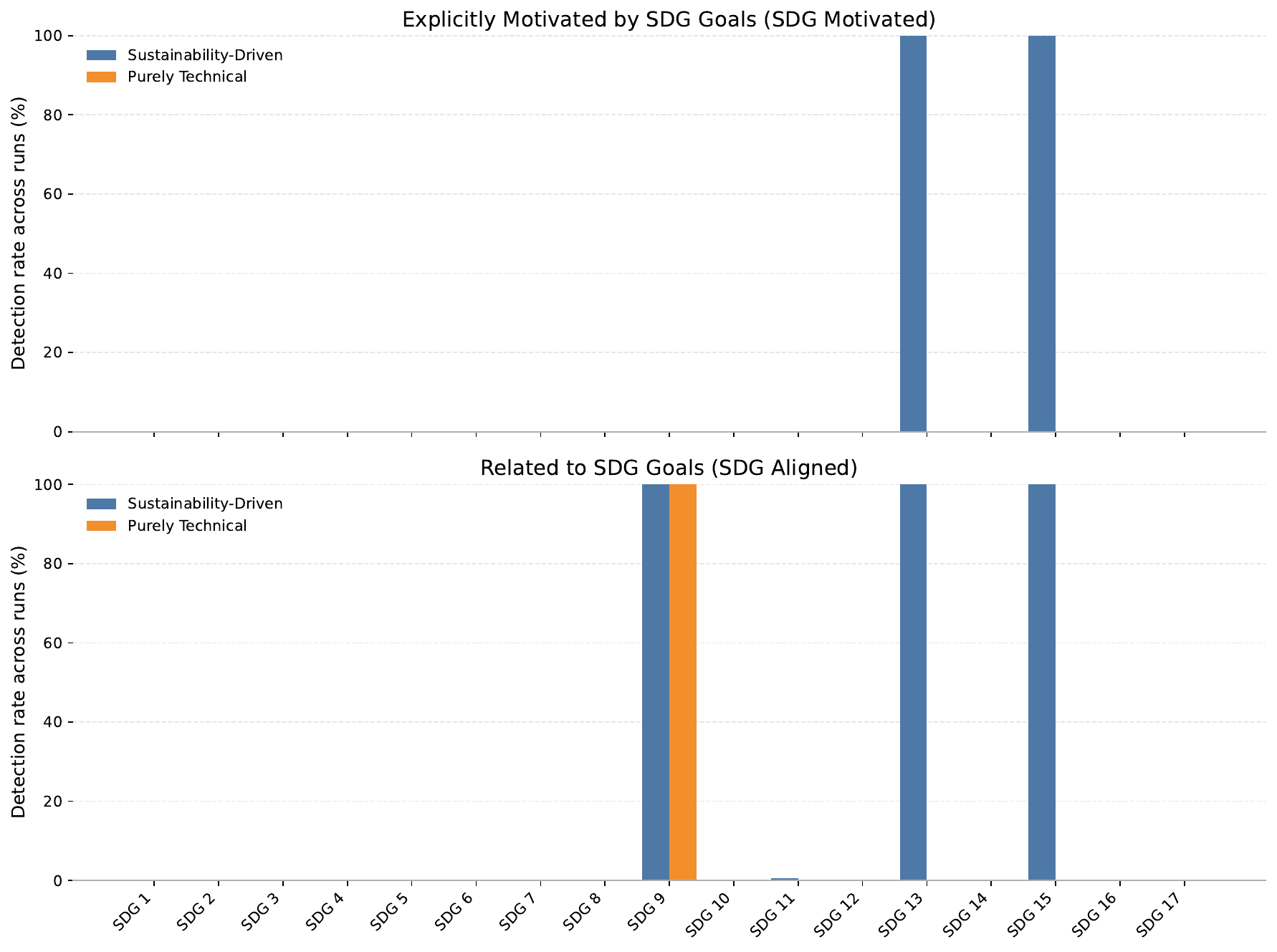}
    \caption{
        Ratio of classified SDG motivation and alignment across 200 runs for both 
        Example A(Purely technical) and Example B (Sustainability-Driven). 
        The above plot shows the distribution of SDGs that the papers were classified 
        as motivated by (SDG Motivated). The bottom plot shows the distribution of 
        SDGs that the papers were classified as relevant or aligned to (SDG Aligned).
    }
    \label{fig:consistency_sdg_rates}
\end{figure}

The results of the SDG motivation and alignment consistency analysis are 
shown in \Cref{fig:consistency_sdg_rates}. The sustainability-driven paper (Example B) is 
consistently classified as explicitly motivated by SDG 15 (Life on Land) and SDG 13 (Climate Action), 
while the purely technical paper (Example A) is consistently classified as not explicitly 
motivated by any SDG. At the same time both papers are consistently classified as aligned to 
SDG 9 (Industry, Innovation and Infrastructure), while the sustainability-driven paper (Example B) is 
also consistently classified as aligned to SDG 15 (Life on Land) and SDG 13 (Climate Action).

\begin{figure}[!htb]
    \centering
    \includegraphics[width=\textwidth]{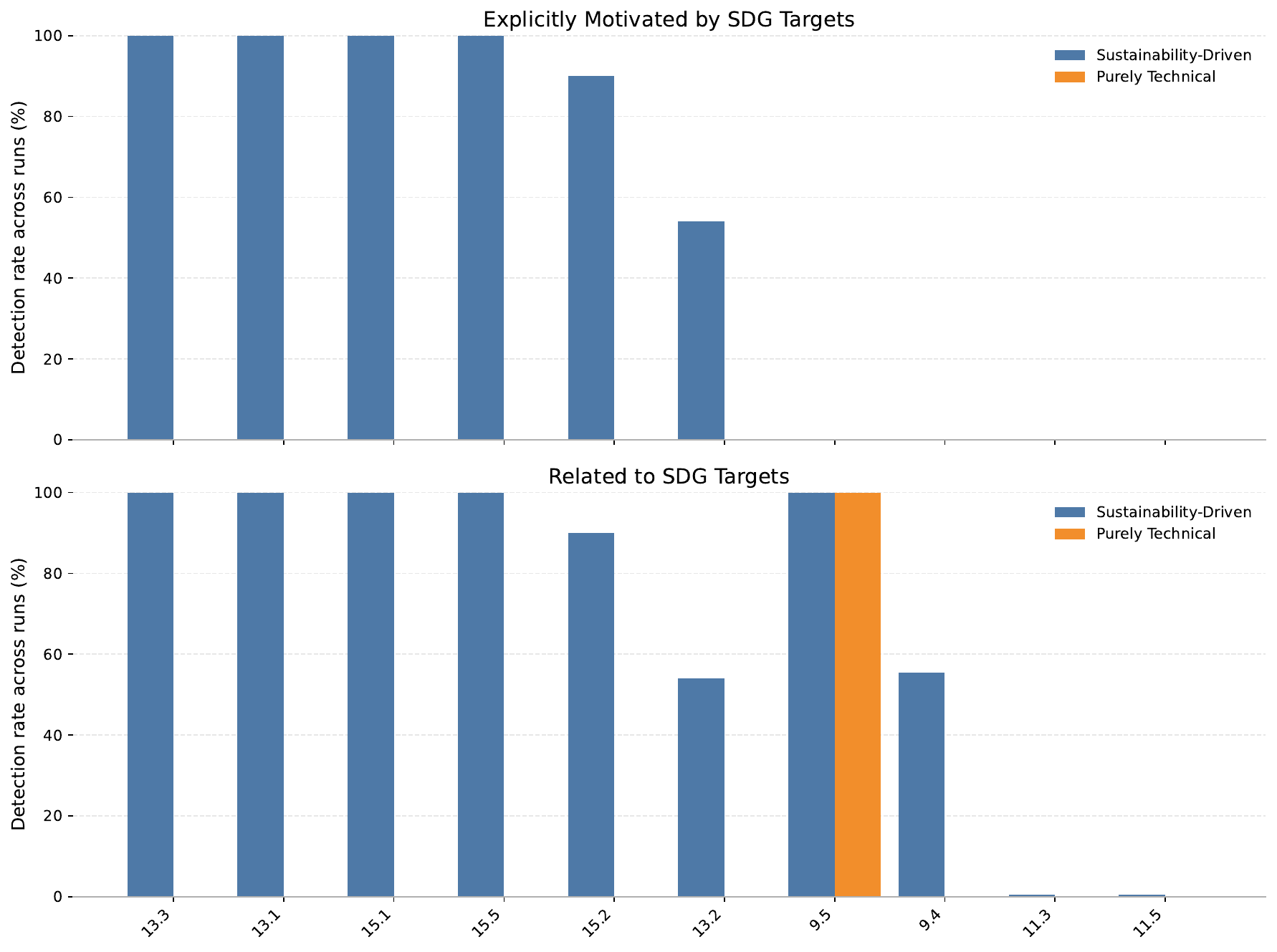}
    \caption{
        Ratio of classified SDG motivation and alignment targets across 200 runs for both 
        Example A(Purely technical) and Example B (Sustainability-Driven). 
        The above plot shows the distribution of SDG targets that the papers were classified 
        as motivated by (SDG Motivated). The bottom plot shows the distribution of 
        SDG targets that the papers were classified as relevant or aligned to (SDG Aligned).
    }
    \label{fig:consistency_sdg_target_rates}
\end{figure}

\Cref{fig:consistency_sdg_target_rates} further demonstrates the consistency of the model's 
classifications at the target level. The sustainability-driven paper (Example B)
is consistently classified as explicitly motivated by targets:
\begin{itemize}
    \item \textbf{13.3} - Improve education, awareness-raising and human and institutional capacity on climate change mitigation, adaptation, impact reduction and early warning.
    \item \textbf{13.1} - Strengthen resilience and adaptive capacity to climate-related hazards and natural disasters in all countries.
    \item \textbf{15.1} - Ensure the conservation, restoration and sustainable use of terrestrial and inland freshwater ecosystems and their services, in particular forests, wetlands, mountains and drylands, in line
    \item \textbf{15.5} - Take urgent and significant action to reduce the degradation of natural habitats, halt the loss of biodiversity and, by 2020, protect and prevent the extinction of threatened species.
\end{itemize}

Furthermore in about 50-90\% of runs the paper is also classified as motivated by 
\textbf{15.2} (Promote the implementation of sustainable management of all types of 
forests, halt deforestation, restore degraded forests and substantially increase 
afforestation and reforestation globally.) While at about 50\% of runs it is additionally 
classified as motivated by \textbf{13.2} (Integrate climate change measures into 
national policies, strategies and planning.) The quotes from the paper justifying 
this classification are consistently:
\begin{verbatim}
    The challenges of climate change and environmental degradation call 
    for innovative solutions that can support sustainable practices...
    Environmental monitoring is one of the crucial domains among the many 
    areas in sustainable robotics. Forest canopies, in particular, are critical 
    for biodiversity, water cycles, and carbon storage.
\end{verbatim}
The sustainability-driven paper (Example B) is also consistently classified a
s aligned to the same above mentioned targets. However it is also classified as relevant to
the SDG 9 targets, in particular to \textbf{9.5} (Enhance scientific research,
 upgrade the technological capabilities of industrial sectors in all countries,
  in particular developing countries, including, by 2030, encouraging innovation 
  and substantially increasing the number of research and development workers per 
  1 million people and public and private research and development spending.) and in about 
50\% of runs to \textbf{9.4} (By 2030, upgrade infrastructure and retrofit 
industries to make them sustainable, with increased resource-use efficiency
 and greater adoption of clean and environmentally sound technologies and 
 industrial processes, with all countries taking action in accordance with
  their respective capabilities.).

Interestingly in 1 of the runs it is also classified as aligned to the SDG 11, targets
\textbf{11.3} (By 2030, enhance inclusive and sustainable urbanization and capacity 
for participatory, integrated and sustainable human settlement planning and 
management in all countries.) and \textbf{11.5} (By 2030, significantly reduce 
the number of deaths and the number of people affected and substantially 
decrease the direct economic losses relative to global gross domestic product 
caused by disasters, including water-related disasters, with a focus on protecting 
the poor and people in vulnerable situations.). The reasoning provided by the model is 
\begin{verbatim}
    The method is proposed for efficient monitoring of forest areas 
    (a type of human settlement/natural community), contributing to sustainable 
    management and disaster risk reduction (e.g. identifying ecological threats)
\end{verbatim}

Rather than an error, this illustrates the model's semantic 
flexibility and its ability to identify valid, although somewhat tangential, 
interdisciplinary connections. It highlights the value of enforcing 
a structured reasoning output (Point 5 in our prompt), which allows 
human reviewers to verify that even minority classifications are 
grounded in sound logic rather than hallucination.

The purely technical paper (Example A) is consistently classified as not 
explicitly motivated by any SDG target. While it it is consistently classified as 
aligned to the SDG 9 target \textbf{9.5}.

\subsection{Footprint of the consistency analysis}
The consistency analysis was conducted on a total of 400 runs (200 runs for each of the two papers). 
The total number of API calls made was 400, with the average prompt size of around 25,000 tokens 
per paper (including the system prompt and the paper-level prompt).

Using the data by \citet{jegham2025hungry}, the total energy consumption of the 
consistency analysis was around $400 \times 13.162Wh = 5.264kWh$, 
and the carbon footprint was around $400 \times 0.013kWh \times 0.6 = 3.12kgCO_2eq$. The overall 
cost of the analysis was around 7\$ (for 400 papers).

\end{document}